\documentclass[10pt]{article}

\PassOptionsToPackage{dvipsnames}{xcolor}
\usepackage[margin=1in]{geometry}
\usepackage[numbers,sort&compress]{natbib}

\usepackage{iftex}
\ifPDFTeX
  \usepackage[utf8]{inputenc}
  \usepackage[T1]{fontenc}
\fi

\usepackage{hyperref}
\usepackage{url}
\usepackage{microtype}
\usepackage{graphicx}
\usepackage{subcaption}
\usepackage{wrapfig}
\usepackage{booktabs}
\usepackage[table]{xcolor}
\usepackage{amsmath}
\usepackage{amssymb}
\usepackage{amsfonts}
\usepackage{mathtools}
\usepackage{amsthm}
\usepackage{nicefrac}
\usepackage{bbm}
\usepackage{bm}
\usepackage{multirow}
\usepackage{tabularx}
\usepackage{enumitem}
\usepackage{placeins}
\usepackage[nameinlink,capitalize]{cleveref}
\crefname{section}{Sec.}{Secs.}
\Crefname{section}{Section}{Sections}
\crefname{subsection}{Sec.}{Secs.}
\Crefname{subsection}{Section}{Sections}
\crefname{figure}{Fig.}{Figs.}
\Crefname{figure}{Figure}{Figures}
\crefname{table}{Tab.}{Tabs.}
\Crefname{table}{Table}{Tables}
\crefname{appendix}{App.}{Apps.}
\Crefname{appendix}{Appendix}{Appendices}
\crefname{definition}{Def.}{Defs.}
\Crefname{definition}{Definition}{Definitions}
\crefname{theorem}{Thm.}{Thms.}
\Crefname{theorem}{Theorem}{Theorems}

\theoremstyle{plain}
\newtheorem{theorem}{Theorem}[section]
\newtheorem{proposition}[theorem]{Proposition}

\theoremstyle{definition}
\newtheorem{definition}{Definition}[section]

\newtheorem{example}[definition]{Example}
\theoremstyle{remark}

\usepackage{physics}
\usepackage{bm}

\def\cD{{\mathcal{D}}}

\def\cL{{\mathcal{L}}}

\def\cP{{\mathcal{P}}}

\def\cT{{\mathcal{T}}}

\def\bE{{\mathbb{E}}}

\DeclareMathOperator*{\argmin}{arg\,min}

\title{Discovering Learning-Friendly Generation Orders\\for Sequential Computation}
\date{}

\author{%
  Yuta Sato$^{1}$, Kazuhiko Kawamoto$^{1}$,
  Hiroshi Kera$^{1,2}$\thanks{Corresponding author: Hiroshi Kera (\texttt{kera@chiba-u.jp}).}\\[0.5em]
  $^{1}$ Chiba University \qquad $^{2}$ National Institute of Informatics
}

\begin{document}

\maketitle

\begin{abstract}
Sequential computation via autoregressive generation can make difficult tasks learnable,
but the generation order of intermediate states strongly affects whether training succeeds.
We address the problem of discovering a learning-friendly target order automatically,
rather than relying on task-specific design.
Our key observation is that learning-friendly orders cause a faster loss drop in the early
stage of training.
We exploit this by \emph{loss profiling}, which ranks candidate orders by the early-stage
loss of a single short run.
To handle the factorial candidate space, we wrap loss profiling in a hierarchical
global--local search over block- and within-block-level orderings.
On six order-sensitive tasks, the method discovers effective orders up to $L=13$ from
random initialization and up to $L=40$ from structured initialization, lifting success
rates from about 10\% to near 100\%.
On integer multiplication, it rediscovers the reverse-digit order that was reported to be
efficient in prior studies.
On delay dynamical systems, as a case study of multi-variate recurrences, learnability
varies sharply even among valid topological sorts of the dependency graph: loss profiling
identifies a learning-friendly one, and the global search even discovers orders surpassing
hand-designed candidates.
\end{abstract}

\section{Introduction} \label{intro}
Discovering computational patterns in arithmetic and symbolic computation is challenging even for deep learning models. For example, learning the parity function---predicting the parity of an input bit string---has been theoretically shown to be hard for gradient-based learning~\citep{failure} and encoder-only Transformer~\citep{Sensitive}. This is because the target functions (e.g., parity) are input-sensitive; a small input change generally alters the output. This contrasts with classical tasks such as image classification, where outputs are expected to be insensitive to random input perturbations. Hence, learning input-sensitive functions is of fundamental interest, and many techniques have been developed to improve empirical learnability towards hard arithmetic and symbolic computation tasks~\citep{LWE,LWE2,LWE3,Grobner,Border_basis,Lyapunov}.

A powerful approach to address this is \textit{sequential computation} via autoregressive generation. A computational task is broken down into subproblems, and the model autoregressively predicts a sequence of intermediate computation tokens towards the final answer. For example, \citet{CoT_parity} has proven that the step-by-step prediction of the parity of the first $k$ bits for $k=1, 2, \ldots$, makes parity function learning successful, as it turns the parity computation of $k+1$ bits into 2-bit parity computation (i.e., $k$-th sub-parity and the ($k+1$)-th bit). As such, a full utilization of the sequential computation makes several almost unlearnable tasks learnable. 

\begin{wrapfigure}{r}{0.38\linewidth}
    \vspace{-1.1em}
    \centering
    \includegraphics[width=\linewidth]{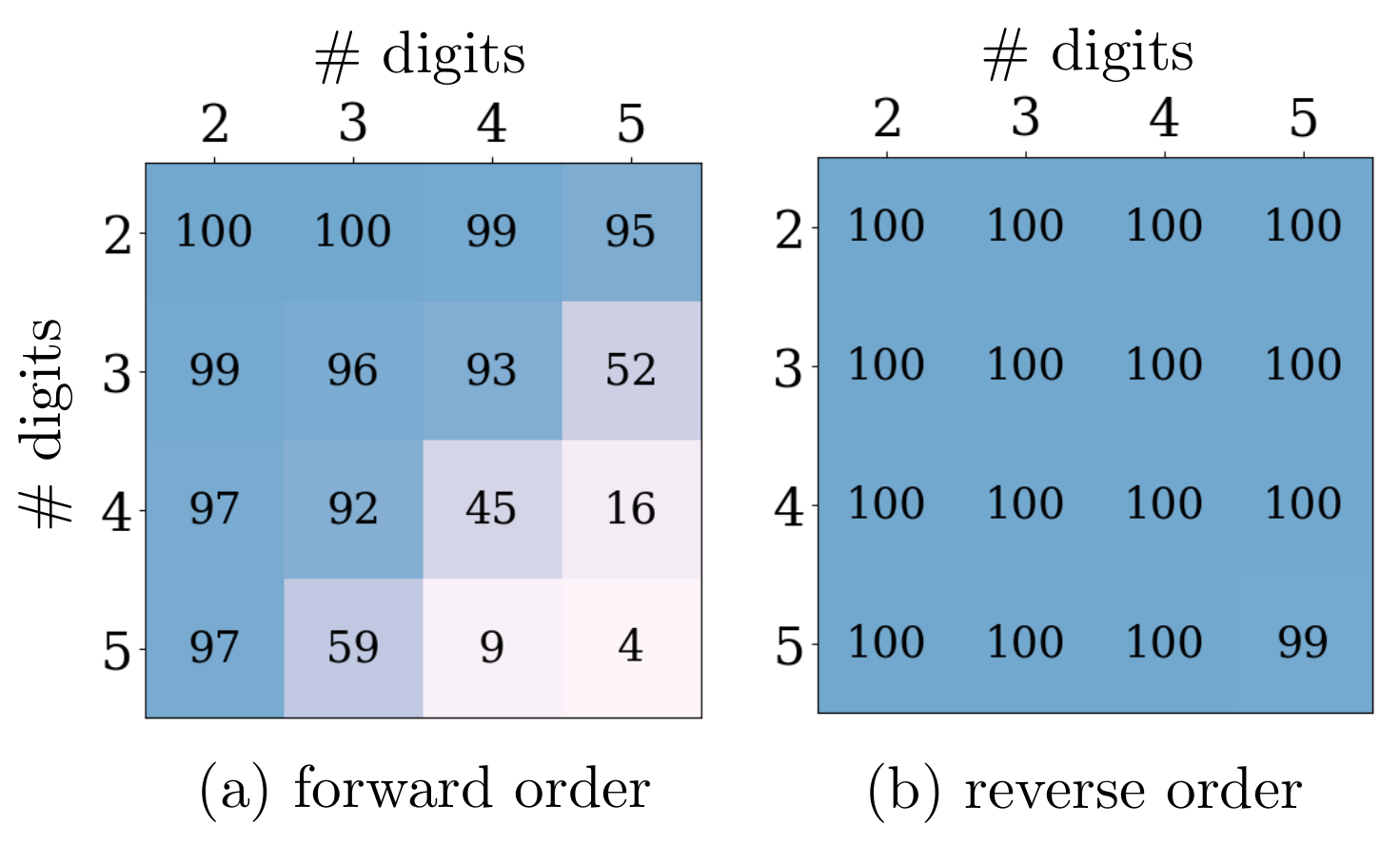}
    \caption{Success rates for integer multiplication from~\citet{Order} (reproduced). The rows/columns indicate the number of digits in the two operands. }
    \label{fig:regular-vs-reverse-in-prod}
    \vspace{-1mm}
  \end{wrapfigure}

An important yet underexplored aspect is the order of sequential computation.
\citet{Order} has shown that Transformers learn multiplication of two integers with better generalization to larger integers when the product is predicted from least to most significant digits~(cf.~\cref{fig:regular-vs-reverse-in-prod}). Concretely, a GPT2-small trained on 300k instances of up-to-five-digit multiplication attains near-perfect accuracy with the least-significant-first order, but performs substantially worse with the most-significant-first order. 
In both cases, the model performs sequential prediction of digits, but the order matters. 
Manual choices of an order rely on task-specific knowledge (e.g., that carries propagate from least to most significant digits in multiplication), and become infeasible when the underlying computational pattern is itself unknown. A systematic procedure for determining a learning-friendly order is therefore necessary.

This study addresses the task of reordering decoder input tokens into a learning-friendly order for better learning of sequential computation tasks. 
Exploiting the observation that neural networks tend to learn from easy to hard instances during training~\citep{Memorization_DNNs, detect_noise, DataMap}, we train a lightweight Transformer on a mixture of target sequences in different orders and identify those that lead to a faster loss drop in the early stages of training. 
To handle longer sequences, we propose a two-stage hierarchical approach, where the global stage finds block-level orders, while the local stage refines the order within each block.

The experiments demonstrate that the proposed method turns training successful by reordering target sequences. 
We designed six order-sensitive tasks that are relatively easy to compute with the (input and) target sequence in the forward order, but not with other orders. The proposed method succeeds in discovering the optimal order for sequences of length up to 13 (i.e., roughly $10^{9}$ permutations) with randomized initial candidates, and structured candidates extend it to length 40 (i.e., roughly $10^{47}$ permutations), increasing the success rate from approximately 10\,\% to 100\,\%. 
We also applied our method to the multiplication task in \citet{Order} and successfully rediscovered the reverse orders.
As a more realistic test, we further apply the pipeline to delay dynamical systems, a class of multi-variable recurrences common in physical, chemical, and biological modeling. Across six variants, the pipeline identifies a topologically interleaved order that substantially outperforms the natural time-major flattening, and the global search even discovers orders that surpass the best hand-designed candidate.

Our contributions are summarized as follows:
\begin{itemize}
    \item We address a novel task of discovering a learning-friendly order of decoder input tokens. We define order-sensitive tasks and introduce six instances of recurrence-based sequential computation that serve as a benchmark for this new task. 
    \item We propose a two-stage exploration method that efficiently determines learning-friendly orders. The exploration maintains a set of candidate permutations, and the candidates are ranked and replaced based on the loss profiling, which exploits the nature of easy-to-hard learning of neural networks.
    \item Our experiments demonstrate the broad applicability of the method. On six order-sensitive tasks, it scales to sequence lengths up to $L=13$ with random initial candidates and up to $L=40$ with structured ones, and it rediscovers the reverse order reported by~\citet{Order} for integer multiplication. On delay dynamical systems, a case study of multi-variable recurrences, learnability varies sharply even among valid topological sorts of the dependency graph; loss profiling identifies a learning-friendly one across multiple variants, and the global search even surpasses hand-designed candidates.
\end{itemize}


\section{Related work}
\vspace{1mm}
\noindent\textbf{Transformers for sequential computation.}
Transformers have been applied to a wide range of sequential computation tasks, including arithmetic~\citep{grokking,length_arithmetic, Abucas, weight_decay}, symbolic computation~\citep{Symbolic_integral, Linear_Algebra, Lyapunov, Grobner, Border_basis}, and number-theoretic problems~\citep{LWE, LWE2, LWE3, gcd, int2int}.
Some of these tasks are known to be difficult to learn end-to-end; for example, parity prediction is provably hard for gradient-based learners~\citep{failure} and for encoder-only Transformers~\citep{Sensitive}.
A remedy is to introduce intermediate computation steps, as~\citet{CoT_parity} have shown for parity by predicting the parities of the first $k$ bits step by step.
Within this line of work, several axes have been identified as critical to learnability, including the choice of token representation and positional encoding~\citep{length_arithmetic, Abucas, weight_decay}, the training distribution and learned rules~\citep{gcd}, and length generalization~\citep{weight_decay, length_arithmetic}.
This study focuses on a less-explored axis: the order in which the intermediate states are emitted by the autoregressive decoder.

\vspace{1mm}
\noindent\textbf{Generation order in autoregressive learning.}
The most relevant prior work is~\citet{Order}, who has shown that Transformers generalize integer multiplication to longer integers when the product is generated from the least significant digit upward.
The reverse-digit ordering was chosen heuristically, based on the observation that carries propagate from least to most significant digits, and the same procedure has not been generalized to other sequential computation tasks where the underlying computational pattern is less transparent.
A related but separate line of work studies generation order in natural language generation.
\citet{brantley-etal-2019-non} train text generators that decide their own generation order by recursively expanding a binary tree, and \citet{where_to_unmask} learn an unmasking order at inference time for masked diffusion language models. 
These methods aim to improve generation quality at inference, assuming that the underlying language model can already be trained successfully under the default order (e.g., left-to-right autoregression or standard masked diffusion).
In contrast, our study focuses on learnable orders at training time; in the sequential computation tasks we consider, most orders make training itself fail or extremely difficult.


\section{Problem Setup}
\label{sec:unraveling-CoT}

Our problem is to discover a target sequence order of the Transformer decoder that improves learning. We first set up our task and then define order sensitivity. Lastly, we clarify the scope of this work.

\subsection{Discovering learning-friendly orders.} 
Let $\Sigma$ be the set of all tokens. We denote the set of all finite token sequences by $\Sigma^*$ and its restriction to length-$L$ sequences by $\Sigma^L$. Let $\cT_{\theta}: \Sigma^* \times \Sigma^L \to \Sigma^L$ be a Transformer with parameter $\theta$. We assume that the target sequence length is fixed. Let $(X, Y) \sim \cD$ be an input--target sequence pair $(X, Y)$ with $|Y| = L$ from a joint distribution $\cD$. Let $S_L$ be the set of all permutations over $\{1,\ldots, L\}$. 
The empirical risk minimizer $\theta_{\mathrm{ERM}}^{\pi}$ with finite sample set $D = \{(X_i, Y_i)\}_{i=1}^{m}$ and permutation $\pi\in S_L$ is 
\begin{align}
    \theta_{\mathrm{ERM}}^{\pi} = \argmin_{\theta} \ \frac{1}{m} \sum_{i=1}^m\ell(\cT_{\theta}, X_i, \pi(Y_i)),
\end{align}
where $\ell$ is a loss function. 
Our goal is to discover a permutation $\pi$ that minimizes the expected risk:
\begin{align}
    \pi^* = \arg\min_{\pi \in S_L}  \bE_{(X, Y) \sim \cD} \qty[ \ell\qty( \cT_{\theta_{\mathrm{ERM}}^{\pi}}, X, \pi(Y)) ].
\end{align}
The choice of $\pi$ is crucial particularly for sequential computation tasks. With a generic order (e.g., random order), the task is almost unlearnable, while an optimal order leads to nearly 100\% success rate~(cf.~\cref{tab:success_rate_each_order}: forward vs. reverse orders). We next discuss such \emph{order sensitive} tasks.

\subsection{Order sensitive tasks}\label{sec:order-sensitive-tasks}
If the operation counts in autoregressive computation for a task have a non-trivial gap between two orders, approaching infinity as the sequence length grows, we refer to the task as order sensitive. We leave the formal definition of order sensitivity to \cref{app:order_sensitive_proofs} due to space constraints, and instead introduce a few examples of order-sensitive tasks to give intuition.


\begin{example}[\textsc{ReLU} sequence]
Let $X = [x_1, \ldots, x_L]$, a sequence of integers. The ReLU sequence $Y = [y_1, \ldots, y_L]$ is: $y_1 = x_1$ and $y_i = \mathrm{ReLU}(x_i + y_{i-1}) = \max(x_i + y_{i-1}, 0)$ for $i = 2, \ldots, L$.
\end{example}
Then, autoregressive generation with the input sequence $X$ and the target sequence in natural causal order, i.e., $Y = [y_1, \ldots, y_L]$, determines the next token $y_i$ locally by the immediate context (e.g., $y_{i-1}$ and $x_i$). 
However, in anti-causal orders (e.g., the reverse order $[y_L, \ldots, y_1]$), this locality is broken. The application of the ReLU function is irreversible when $x_i + y_{i-1} \le 0$, and this ambiguity forces the model to implicitly resolve the entire causal chain from the input to determine the correct token, significantly increasing the computational burden compared to the forward order. 

One can define order-sensitive tasks by recurrence and a non-invertible function $f$, i.e., $x_1=y_1$ and $y_i = f(x_i, y_{i-1})$ for $i \ge 2$. \Cref{tab:success_rate_each_order} provides several examples including the \textsc{ReLU} task and success rates over evaluation sets. Their order sensitivity is shown in~\cref{app:order_sensitive_proofs}. These examples will be used as a benchmark for our method. The last three tasks (\textsc{MLP}, \textsc{Sine}, and \textsc{Cubic}) have strong nonlinearity and are introduced as particularly challenging tasks.

\begin{table}
\centering
\small
\setlength{\tabcolsep}{4pt}
\caption{Six order-sensitive tasks and success rates (\%) at $L{=}20$ for Transformers trained with a fixed forward or reverse order; the forward order is substantially more learning friendly across all tasks. $L{=}50$ results in \cref{tab:success_rate_each_order_full}; setup details in \cref{sec:training_setup,app:dataset_examples}.}
\label{tab:success_rate_each_order}
\begin{tabular}{llcc}
\toprule
\textbf{Task} & \textbf{Recurrence ($i \ge 2$)} & \textbf{Forward} & \textbf{Reverse}\\
\midrule
\textsc{ReLU}     & $y_i = \operatorname{ReLU}(x_i + y_{i-1})$                                & \textbf{99.6} & 5.6 \\
\textsc{Square}   & $y_i = (x_i^2 + y_{i-1}^2)\bmod N$                                        & \textbf{100}  & 0.1 \\
\textsc{Triangle} & $y_i = \bigl|((y_{i-1}+x_i)\bmod 2N) - N\bigr|$                           & \textbf{100}  & 0.2 \\
\textsc{MLP}      & $y_i = \bigl\lfloor \operatorname{MLP}(x_i, y_{i-1}) \bigr\rfloor \bmod N$ & \textbf{100}  & 9.4 \\
\textsc{Sine}     & $y_i = \bigl\lfloor A\sin\!\bigl(\tfrac{2\pi}{P}(y_{i-1}+x_i)\bigr) \bigr\rfloor \bmod N$ & \textbf{100}  & 6.2 \\
\textsc{Cubic}    & $y_i = (y_{i-1}^3 + x_i)\bmod N$                                          & \textbf{100}  & 1.6 \\
\bottomrule
\end{tabular}
\end{table}


\vspace{1mm}
\noindent\textbf{Failure of simple alternatives.} Two natural alternatives fail in this setting. (i) Jointly optimizing a soft permutation $\tilde{P} \in [0,1]^{L \times L}$ together with model parameters (target preprocessed as $Y\tilde{P}$) leaks information across positions and undermines next-token prediction (\cref{app:soft_perm_leakage,app:analysis_attn_sparsity}). (ii) An evolutionary-strategy baseline for permutation search did not yield a practical alternative (\cref{app:permutation_search_ES}).

\paragraph{Scope of this work.}\label{sec:scope} We target sequential computation tasks where a model generates structured intermediate states step by step and the generation order strongly affects whether training succeeds.
Their target functions are input-sensitive: small input changes alter the output substantially, leaving little room for shortcut predictions, and the optimal order can be characterized analytically.
Natural language generation is qualitatively different and is not a target of the current work.
Common words have strong co-occurrence patterns (e.g., ``live'' and ``in'') that allow shortcut learning regardless of the order, and even sentence- or paragraph-level reorderings leave function words such as ``the'' or ``it'' that hint at the original order.
Moreover, language models retain moderate accuracy under suboptimal orders, whereas accuracy on order-sensitive tasks collapses to near zero (\cref{tab:success_rate_each_order}).
The proposed method itself is task-agnostic by design, requiring only candidate orders and a loss signal. Extending it to natural-language reasoning is a promising future direction that broadens its generality.

\section{Proposed Method}\label{sec:proposed-method}
We now present a method for discovering learning-friendly orders.
It consists of two components, loss profiling and hierarchical exploration, and is based on the following assumptions.

\vspace{1mm}
\noindent\textbf{Assumption 1: Learning prefers computationally efficient orders.}
For order-sensitive sequential computation tasks, we hypothesize that Transformer training is biased toward orders that require fewer operations.
This is consistent with the empirical tendency of deep models to fit simpler patterns first, which can sometimes lead to shortcut learning.

\vspace{1mm}
\noindent\textbf{Assumption 2: Weak locality property.}
We assume that small local reorderings do not drastically change the learning quality of an order.
In the \textsc{ReLU} recurrence, for example, swapping neighboring targets around the forward order usually introduces only a one-step skip in sequential computation, so the required function class expands only mildly.
In contrast, moving tokens corresponding to late-stage outputs toward the beginning breaks the coarse causal layout, because partial computations that should appear early are pushed to later positions.

An interesting question on Assumption 1 may be whether all computationally efficient orders are learning-friendly. We will show a counter-example in \cref{sec:exp-delay-variants}. Assumption 2 introduces a reasonable restriction on our task; without this assumption, tiny local swaps can abruptly flip learnability, and exhaustive search over full permutations becomes necessary.

\subsection{Loss Profiling}
We first present \emph{loss profiling}. Given a set of candidate permutations, loss profiling narrows down the set to a few learning-friendly orders through short training runs of a small Transformer model on a mixed dataset of variously ordered target sequences. 

\begin{wrapfigure}{r}{0.3\columnwidth}
    \centering
    \vspace{-2mm}
    \includegraphics[width=0.95\linewidth]{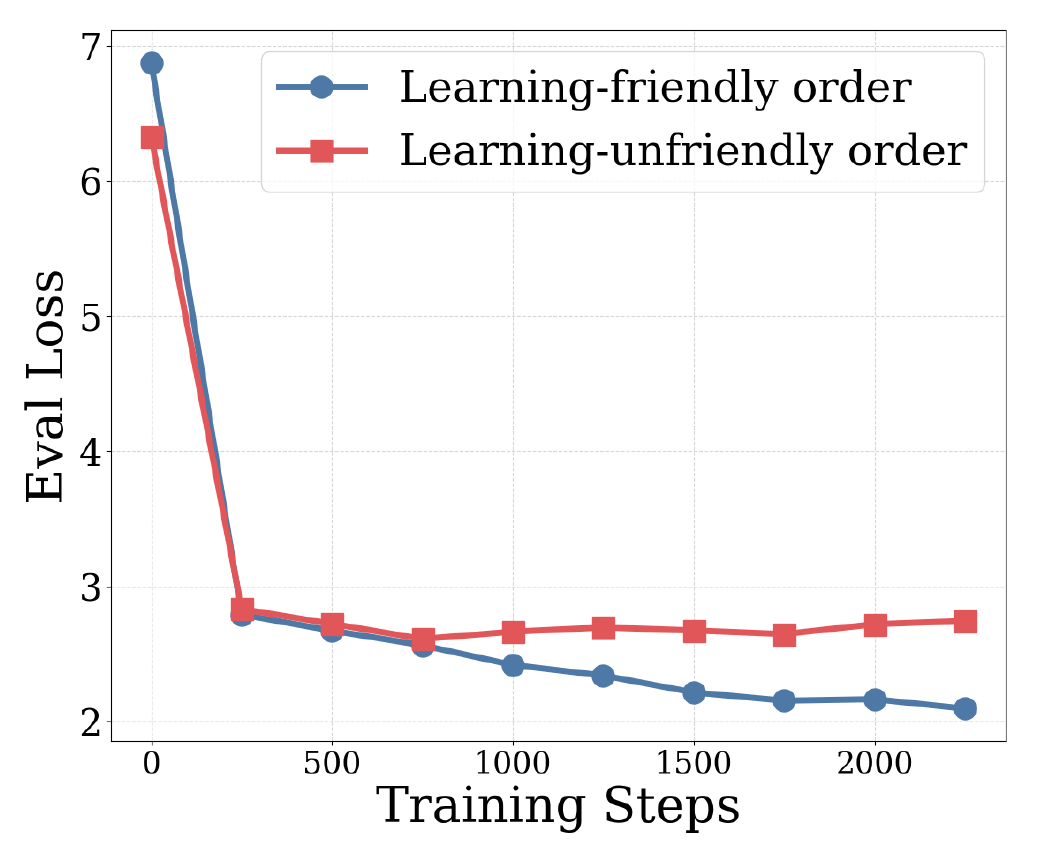}
    \caption{Validation loss curves at early-stage training.}
    \label{fig:loss-profiling-two-order}
    \vspace{-3mm}
\end{wrapfigure}
The key observation is that the validation loss drops faster for learning-friendly orders (\cref{fig:loss-profiling-two-order}). 
This derives from the characteristic of the training dynamics of deep neural networks: they tend to learn easy samples in the early stages of training, and gradually adapt to harder samples later. Several studies have already leveraged this characteristic, such as~\citet{Memorization_DNNs} for learning with noisy labels and~\citet{Lens_difficulty} for identifying difficult examples. 

In loss profiling, a Transformer is trained only for a few epochs on a dataset with various orders in mixture; it then identifies learning-friendly orders as ``easy samples,'' for which the loss drops faster. 

More formally, let $D=\{(X_i,Y_i)\}_{i=1}^{m}$ and $D'=\{(X_i',Y_i')\}_{i=1}^{m'}$ be training and validation sets, respectively.
Let $\mathcal{P}=\{P_1,\dots,P_{T}\}$ be the set of $T$ candidate permutation matrices.  Let $D^{P_t}$ be the set $D$ with reordered target sequences by $P_t$, i.e., $D^{P_t} = \{(X_i, Y_iP_t)\}_{i=1}^m$. The learning-friendly orders are determined as follows. 
\begin{enumerate}[label=\textbf{P\arabic*.}, leftmargin=2em, labelsep=0.4em, itemsep=0.6mm, topsep=0pt]
    \item Train a Transformer for $n_{\mathrm{e}}$ epochs on a mixed dataset $\bar{D} := \bigcup_{t=1}^T D^{P_t}$. Let $\cT_{\theta'}$ be the Transformer after training.
    \item Compute the validation loss for each $P_t \in \cP$ as 
        $\mathcal{L}(D', P_t) = \frac{1}{m^{\prime}} \sum_{i=1}^{m^{\prime}}\ell(\cT_{\theta'}, X'_i, Y'_iP_{t})$.
    Then, output $P^* = \argmin_{P_t\in\cP} \cL(D', P_t)$.
\end{enumerate}

We empirically observed that loss profiling can handle a few thousand permutations. However, the number of permutations grows factorially in the sequence length. Next, we introduce two-stage hierarchical optimization, where the loss profiling is called as a subroutine. 

\subsection{Hierarchical Exploration}

We present the global--local pipeline (cf.~\cref{fig:HPS_flow}). We start with the initial set of permutation candidates $\cP_0 = \{P_1, \ldots, P_T\}$. The global stage splits each token sequence into several blocks and finds a good block-level permutation. 
The local stage refines this coarse ordering by permuting the tokens within each block discovered at the global stage.
Formally, the two stages operate as follows.

\vspace{1mm}
\noindent\textbf{Global stage.}
Let the search depth be $K$ and $T=2K!$. Let $\cP_1 := \cP_0$. For $k=1,\ldots, K$, we split each target sequence into $k$ blocks, and apply loss profiling to the new permutation set: 
\begin{align}\label{eq:block-level-perm}
    \bigcup_{P \in \cP_k} \{PQ_{1}, \ldots, PQ_{k!}\} \cup \{PQ_{1}J, \ldots, PQ_{k!}J\},
\end{align}
where $Q_{i} \in [0, 1]^{L\times L}$ are the block-level permutations and $J \in [0, 1]^{L \times L}$ is the exchange matrix (reversal permutation). 
The best $\lfloor T/2k! \rfloor$ permutations define the new candidate set $\cP_{k+1}$, where $\lfloor \cdot \rfloor$ denotes the floor function. 
We denote the best permutation obtained in the final iteration ($k=K$) as $P_{\mathrm{g}}$.
This permutation is then refined with the local stage.

\vspace{1mm}
\noindent\textbf{Local stage.}
Let $P_1 := P_\mathrm{g}$ be the initial permutation. 
We again conceptually split each target sequence into blocks of size $l$. 
Let $R_{1}^{i}, \ldots, R_{l!}^i \in [0,1]^{L\times L}$ be all the permutations inside the $i$-th block. These permutations do not change the orders outside the $i$-th block. 
For each block length $l \in \{ 2,3,\dots, \lfloor L/2\rfloor\}$\footnote{When $l$ does not divide $L$, the remaining $L\ \text{mod}\ l$ tokens are placed in an additional block.}, we apply loss profiling to the new candidate set:
\begin{align}\label{eq:inter-block-perm}
    \bigcup_{i=1}^{\lceil L/l \rceil}\{PR_{1}^{i}, \ldots, PR_{l!}^{i}\},
\end{align}
and denote the lowest-loss result by $P_l$, where $\lceil \cdot \rceil$ is the ceiling function.
Keeping each block’s internal order fixed, we perform loss profiling over the $\lfloor L/l\rfloor$ block-reordering candidates:
\begin{align}
    \{P_l Q_{1}^{\prime},\,\ldots,\,P_l Q_{\lfloor L/l\rfloor!}^{\prime}\},
\end{align}
where $Q_{i}^{\prime} \in [0, 1]^{L\times L}$ are the block-level permutations. 
The best candidate becomes the initial permutation for the next block size $l+1$.


\vspace{1mm}
\noindent\textbf{Computational overheads.}
Three aspects keep the framework efficient. (i) Each profiling run trains for only 800--1,600 steps (1--2 epochs with $10^{5}$ samples and batch size $128$), since loss-drop differences emerge early. (ii) A single run handles up to $2\times 7! = 10{,}080$ permutations. (iii) The global--local pipeline needs $K$ runs at the global stage and $2(\lfloor L/2 \rfloor - 1)$ at the local stage. Small Transformers suffice in exploration since learning-friendly orders should be model-size independent.

\section{Experiments}\label{sec:experiments}

\subsection{Experimental setup} \label{sec:training_setup}
We provide a basic setup. Further details are given in later sections and in the appendix.

\vspace{1mm}
\noindent\textbf{Datasets.}
We generated datasets for the six tasks given in~\cref{sec:order-sensitive-tasks}.
We considered both fixed-length and variable-length cases.
We used $L \in \{5,6,\ldots,50\}$ for the former, and $L$ from two ranges, $L \in \{20, 21, \ldots,30\}$ and $L \in \{40, 41, \ldots,50\}$, for the latter.
We prepared training (size $10^5$), validation (size $10^3$) for exploration, and evaluation (size $10^3$) sets for final training, using random seeds 42, 84, and 123, respectively.

\vspace{1mm}
\noindent\textbf{Exploration setup.}
The proposed global--local method explores learning-friendly orders. As described in \cref{sec:proposed-method}, small, lightweight GPT-2 models~\citep{GPT2} are repeatedly trained for a small number of steps, and the loss profiling ranks permutation candidates using the validation set. Training settings are detailed in~\cref{app:experimental_settings}. 
The configuration of GPT-2 models and the initialization of the candidate permutations of global--local search will be specified in the next sections.

\vspace{1mm}
\noindent\textbf{Final Training setup.}
\label{sec:pipeline} The proposed global--local exploration returns a learning-friendly order. In the final training phase, a full training of a GPT-2 model is performed on the training set with target sequences in the discovered order, and the success rate over the evaluation set is measured. The success rate, also called sequence-level exact-match accuracy, is defined as the proportion of evaluation samples where the Transformer outputs an entirely correct target sequence. Throughout the paper, ``success rate'' refers to this generation-level metric; the order-discovery performance of loss profiling is reported separately as top-$k$ identification accuracy or as Spearman rank correlation.
The final training uses a larger GPT-2 model than those used in the exploration phase. Particularly, it has six layers and eight attention heads with embedding and feed-forward dimensions $(d_{\mathrm{emb}}, d_{\mathrm{ffn}})=(512,2048)$.

\subsection{Loss profiling for discovering the forward order}\label{sec:exp-loss-profiling}
\begin{figure*}
  \centering
  \captionsetup{belowskip=-4pt}
  \includegraphics[width=1.0\linewidth]{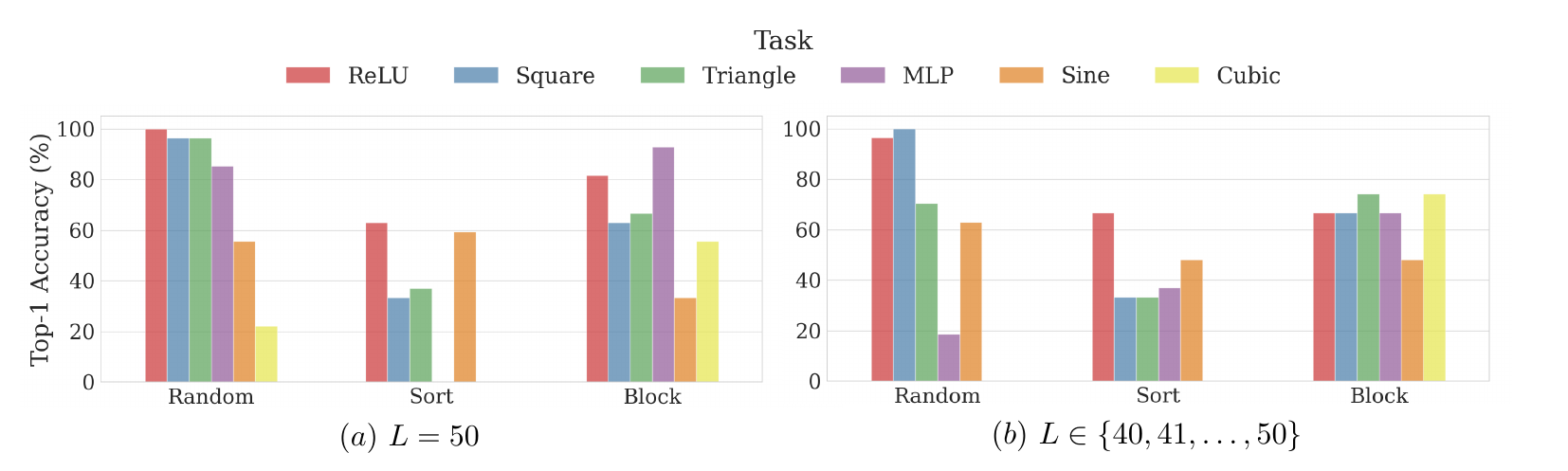}
  \caption{Top-1 identification accuracy of loss profiling:
  (a) fixed-length targets;
  (b) variable-length targets.
  Each run is initialized with one of the permutation sets
  Random $\cP_{\mathrm{r}}$, Sort $\cP_{\mathrm{s}}$, or Block $\cP_{\mathrm{b}}$.}
  \label{fig:loss-profiling-results}
\end{figure*}

We first show that loss profiling successfully discovers the learning-friendly order (i.e., forward order) from candidate sets. We consider three types of candidate sets. \Cref{fig:permutation_sets_with_forward} provides the visualization.

\begin{itemize}[nosep, leftmargin=1.4em, labelsep=0.4em, itemsep=2mm]
    \item $\cP_{\mathrm{r}}$ (\textbf{Random}) consists of the forward order and random orders (permutations). The forward order typically outperforms others by a large margin.
    \item $\cP_{\mathrm{s}}$ (\textbf{Sort}) consists of partially sorted orders. The bubble sort algorithm~\citep{knuth1998art} reorders a random permutation toward the forward order, and we evenly sample permutations from the sorting trajectory. This set ranges from learning-friendly (nearly sorted) to unfriendly (random) orders.
    \item $\cP_{\mathrm{b}}$ (\textbf{Block}) consists of orders that are randomly shuffled at the block level. We fix the block size to $b=5$. The last block can be smaller than $b$ if $b$ does not divide the sequence length. This candidate set is reasonable when one can assume good prior on the local structure.
\end{itemize}


\vspace{1mm}
\noindent\textbf{Setup.}
To show the robustness of loss profiling up to the choice of model configurations, we tested 27 GPT-2 configurations by varying the number of layers $n_{\mathrm{layer}} \in \{1,2,4\}$, attention heads $n_{\mathrm{head}} \in \{1,2,4\}$, and model dimensions $(d_{\mathrm{emb}}, d_{\mathrm{ffn}}) \in \{(128,512), (256,1024), (512,2048)\}$. 
Given a set of permutations $\cP$, loss profiling assigns a score to each permutation.
Top-$k$ accuracy measures whether the forward order appears among the top $k$, with emphasis on $k=1$.
\Cref{app:profiling_retrain_correlation} analyzes whether the full profiling ranking correlates with retraining accuracy over all candidates.

\vspace{1mm}
\noindent\textbf{Results.} In the fixed-length setting (\cref{fig:loss-profiling-results}(a)), $\cP_{\mathrm{r}}$ achieves the highest top-1 accuracy, identifying the ground truth in nearly 90\% of cases across \textsc{ReLU}, \textsc{Square}, \textsc{Triangle}, and \textsc{MLP}, regardless of model size. $\cP_{\mathrm{b}}$ and $\cP_{\mathrm{s}}$ are weaker because they contain candidates structurally similar to the forward order, yielding comparable early-stage loss decreases. The variable-length case (\cref{fig:loss-profiling-results}(b)) gives comparable accuracy, showing that simple padding suffices. A model-size ablation is in \cref{app:loss_profiling_dim_ablation}. Per-task detailed profiling rankings are reported in \cref{app:all_loss_profiling_results}.


\subsection{Global--local exploration of learning-friendly orders}\label{sec:exp-global-local}
We now demonstrate that the proposed method can discover learning-friendly orders.

\vspace{1mm}
\noindent\textbf{Setup.}
The basic setup follows the previous sections. 
We execute the global--local exploration with search depth $K$. We set $K=6$ for short sequences ($10 \le L \le 13$) and $K=7$ for longer ones ($20 \le L \le 40$) for better convergence. The exploration starts with either of the following initial candidate sets: $\mathcal{P}_{\mathrm{r}}^-, \mathcal{P}_{\mathrm{s}}^-, \mathcal{P}_{\mathrm{b}}^-$~(visualized in \cref{fig:permutation_sets_without_forward}). The superscript `-' indicates that the forward order is replaced by a random permutation for $\mathcal{P}_{\mathrm{r}}$ and a block-level random one for $\mathcal{P}_{\mathrm{b}}$. For $\mathcal{P}_{\mathrm{s}}^-$, a random permutation is sorted toward the \textit{reverse} order.
For $L\ge 20$, only the global stage is conducted as the local stage is too costly.
We tested four parameter combinations ($n_{\mathrm{layer}} \in \{1, 2\}, n_{\mathrm{head}} \in \{1, 2\}$) and present the results from the one with the highest success rate after final training. 
Given the analysis in~\cref{fig:loss-profiling-acc-by-dim}, we set $(d_{\mathrm{emb}}, d_{\mathrm{ffn}})$ to $(128, 512)$ for $\mathcal{P}_{\mathrm{r}}$ and $\mathcal{P}_{\mathrm{s}}$, and to $(512, 2048)$ for $\mathcal{P}_{\mathrm{b}}$.

\begin{table*}[t]
\small
\centering
\caption{Success rates of final training using the permutations discovered by the global--local pipeline on the synthetic tasks. \textbf{Bold} indicates that the optimal (forward) order is successfully discovered.}
\label{tab:global-local-main-results}
\begin{tabular}{clcccccc}
\toprule
\multirow{2}{*}{\textbf{Initialization}} & \multirow{2}{*}{\textbf{Target Length}} & \multicolumn{6}{c}{\textbf{Task}} \\
\cmidrule(lr){3-8}
 & & \textsc{ReLU} & \textsc{Square} & \textsc{Triangle} & \textsc{MLP} & \textsc{Sine} & \textsc{Cubic} \\
\midrule
\multirow{3}{*}{$\cP_{\mathrm{r}}^-$}
 & $L=10$ &  \textbf{99.6} & \textbf{100.0} & \textbf{100.0} & 38.2 & 54.2 & 0.3  \\
 & $L=13$ & \textbf{99.2} & 5.6 & 7.4 & 10.8 & 26.3 & 0.2 \\
 & $L=20$ & 24.2 & 10.4 & 0.8 & 5.9 & 8.2 & 0.0 \\
 \midrule
\multirow{3}{*}{$\cP_{\mathrm{s}}^-$} 
 & $L=20$ & \textbf{99.6} & \textbf{100.0} & \textbf{100.0} & 89.4 & \textbf{100.0} & 1.9 \\
 & $L=30$ & \textbf{97.6} & \textbf{100.0} & \textbf{100.0} & \textbf{100.0} & \textbf{100.0} & 3.8 \\
 & $L=40$ & 3.4 & 0.0 & 0.0 & 0.4 & 1.2 & 0.0 \\
\midrule
\multirow{3}{*}{$\cP_{\mathrm{b}}^-$} 
 & $L=20$ & \textbf{99.6} & \textbf{100.0} & \textbf{100.0} & \textbf{100.0} & \textbf{100.0} & \textbf{100.0} \\
 & $L=30$ & \textbf{97.6} & \textbf{100.0} & \textbf{100.0} & \textbf{100.0} & 1.7 & 0.7 \\
 & $L=40$ & \textbf{96.0} & 2.1 & 1.2 & \textbf{100.0} & 0.0 & 0.0 \\
\bottomrule
\end{tabular}
\end{table*}

\vspace{1mm}
\noindent\textbf{Results.}
With $\cP_{\mathrm{r}}^-$, the method singles out the forward order from $13! > 6\!\times\!10^9$ permutations for \textsc{ReLU} ($L=13$) and from $10! > 3\!\times\!10^6$ for \textsc{Square} and \textsc{Triangle} ($L=10$); structured initializations $\cP_{\mathrm{s}}^-$ and $\cP_{\mathrm{b}}^-$ extend this to larger $L$, with $\cP_{\mathrm{b}}^-$ most powerful---reflecting the utility of local-order priors. On the harder \textsc{MLP}, \textsc{Sine}, and \textsc{Cubic} tasks, structured initializations bring substantial gains for \textsc{MLP} and \textsc{Sine}; \textsc{Cubic} remains the hardest: the cubic-modulo mapping is difficult to learn even with the optimal order. The varying difficulty makes \textsc{MLP}, \textsc{Sine}, and \textsc{Cubic} useful benchmarks for future methods. Per-stage runtimes are in \cref{tab:runtime_analysis_appendix}. Ablation of the local stage, the reverse-order baseline at every length, and the detailed discovered orders appear in \cref{app:global_local_ablation}.

\subsection{Rediscovering the optimal order for the \textsc{Prod} task}
\label{sec:exp-prod}

We next tackle the \textsc{Prod} task to retrieve the reverse order reported in \citet{Order}.

\vspace{1mm}
\noindent\textbf{\textsc{Prod}.} 
Let $L$ be even. 
The input sequence is a concatenation of two left-zero-padded digits, i.e., $X = [a_1, \ldots, a_{L/2}, b_1, \ldots, b_{L/2}]$, where $a_k$, $b_k$ are the $k$-th significant digits of integers $a$ and $b$, respectively. 
The target sequence $[c_1, \ldots, c_L]$ is the left-zero-padded sequence of digits of $ab$. As \cref{fig:regular-vs-reverse-in-prod} shows, the reverse order is known to be learning-friendly, while the forward order is not.


\vspace{1mm}
\noindent\textbf{Setup.}
The general setup follows the previous experiments. Because multiplication involves complex carry propagation, the final training uses a larger GPT-2 ($n_{\mathrm{layer}}{=}12$, $n_{\mathrm{head}}{=}12$, $(d_{\mathrm{emb}}, d_{\mathrm{ffn}}){=}(768, 3072)$, following \citep{Order}) and an enlarged training set of $10^6$ samples. The digit counts of $a$ and $b$ are sampled uniformly from $\{1,\dots,L/2\}$ for training and from $\{L/2{-}2, L/2{-}1, L/2\}$ for evaluation, where hard cases concentrate (\cref{fig:regular-vs-reverse-in-prod}).

\begin{table}[t]
\small
\centering
\setlength{\tabcolsep}{4pt}
\begin{minipage}[t]{0.58\textwidth}
\centering
\caption{Success rates (\%) on \textsc{Prod} using the discovered orders; \textbf{bold} marks runs that recovered the reverse order.}
\label{tab:prod_results}
\vspace{0.3em}
\begin{tabular}{clcc}
\toprule
\multirow{2}{*}{\textbf{Initialization}} & \multirow{2}{*}{\textbf{Target Length}} & \multicolumn{2}{c}{\textbf{Success Rate (\%)}} \\
\cmidrule(lr){3-4}
 & & \textbf{Discovered} & \textbf{Forward} \\
\midrule
\multirow{2}{*}{$\mathcal{P}_{\mathrm{r}}^-$}
  & $L=10$ & \textbf{100.0} & 72.9 \\
  & $L=12$ & 51.4 & 51.4 \\
\midrule
$\mathcal{P}_{\mathrm{s}}$  & $L=20$ & \textbf{98.2} & 9.7 \\
\midrule
$\mathcal{P}_{\mathrm{b}}^-$ & $L=20$ & \textbf{98.2} & 9.7 \\
\bottomrule
\end{tabular}
\end{minipage}
\hfill
\begin{minipage}[t]{0.40\textwidth}
\centering
\caption{Runtime breakdown (in minutes) of the exploration and final training phases for the \textsc{Prod} task.}
\label{tab:runtime_prod}
\vspace{0.3em}
\begin{tabular}{crrr}
\toprule
\multirow{2}{*}{\textbf{Params}} & \multicolumn{2}{c}{\textbf{Exploration}} & \textbf{Final} \\
\cmidrule(lr){2-3}
 & \textbf{Global} & \textbf{Local} & \textbf{Training} \\
\midrule
$K=6, L=10$ & 116.7 & 78.8  & 470.2 \\
$K=6, L=12$ & 121.7 & 102.2 & 511.6 \\
$K=7, L=20$ & 599.7 & ---   & 639.5 \\
\bottomrule
\end{tabular}
\end{minipage}
\end{table}


\vspace{1mm}
\noindent\textbf{Results.}
\Cref{tab:prod_results} shows successful rediscovery of the reverse order for $\cP_{\mathrm{r}}^-$ at $L=10$ and for $\cP_{\mathrm{s}}, \cP_{\mathrm{b}}^-$ at $L=20$. The search fails at larger $L$; learning is too difficult even with the reverse order, so it loses its early-stage edge over other orders. \Cref{tab:runtime_prod} shows that exploration is shorter than even a single full training run; exhaustive order ranking by retraining would require many full runs and is impractical.

\subsection{Delay dynamical systems: a case study}\label{sec:exp-delay-variants}
Previous sections examined order-sensitive one-dimensional recurrences. As a case study of multi-variate recurrence, we now consider delay dynamical systems, where variables depend on lagged values of themselves/others. Delays naturally arise, e.g., from distinct biochemical reaction rates~\citep{erneux2009applied}.

\begin{example}[ReLU delay network]
ReLU delay network is a four-variable delayed dynamical system with state $s(t)=(a(t), b(t), c(t), d(t))$. Two initial states $s(1), s(2)$ are sampled uniformly from $[-25, 25]^2$; the subsequent states are defined by the recurrence
$a(t{+}2) = \operatorname{ReLU}(b(t{+}1)+d(t))$,
$b(t{+}2) = a(t{+}1)-c(t{+}1)$,
$c(t{+}2) = b(t)+d(t{+}1)$,
$d(t{+}2) = \operatorname{ReLU}(a(t{+}1)+c(t))$.
We fix the number of time steps to $T=11$, so the input is the eight integers from $s(1), s(2)$ and the target is the time-major flattening of $s(3),\ldots,s(11)$, a length-$36$ sequence.
\end{example}

\subsubsection{Loss profiling on topological sorts.}
In applications, network topology is sometimes known. For example, the dependency structure of gene regulatory networks can often be inferred from gene knock-out, yet interaction strengths and nonlinearities remain unknown. This motivates selecting a learning-friendly output order from topological sorts, where variables of different time steps are sorted in accordance with the dependency, for time-series prediction.
We compare five target orders (see \cref{tab:delay-variants-summary}). 
The first four orders are valid topological sorts of the dependency graph, while the last one, \texttt{reverse}, is a worst-case baseline. 

\vspace{1mm}
\noindent\textbf{Setup.}
We run loss profiling with a 1-layer GPT-2 for 1 epoch to rank the five candidates, then retrain a 6-layer GPT-2 for 10 epochs with each order fixed and evaluate success rate on a held-out test set (\cref{sec:pipeline}); training/test sets contain $100{,}000/1{,}000$ samples. Besides the \textsc{ReLU} delay network defined above, we use five more delay networks:
\textsc{Hidden-cause} introduces a latent variable $h(t)$ that mediates the updates of $a$ and $d$;
\textsc{Coupled ring} cyclically couples each variable to its neighbor with ReLU subtractions;
\textsc{Modular} uses additive and multiplicative coupling under modulo-$19$ arithmetic;
\textsc{Sine} replaces ReLU with a bounded sinusoidal nonlinearity;
\textsc{Mixed nonlinear} combines ReLU, modular, and squared coupling within the same recurrence.
See~\cref{app:delay-variants-defs} for details.

\begin{table}[t]
\centering
\small
\setlength{\tabcolsep}{4pt}
\caption{Success rate (\%) on six delay networks with $T=11$, after retraining a GPT-2 for 10 epochs with each order. \textbf{Bold} entries mark the order ranked first by loss profiling. Spearman $\rho$ is computed between profile ranks and retrain success rates over the five candidates. See \cref{app:delay-variants-defs} for full definitions.}
\label{tab:delay-variants-summary}
\begin{tabular}{lcccccc}
\toprule
\multirow{3}{*}{\textbf{Task}} & \texttt{time\_first} & \texttt{var\_first\_topo} & \texttt{dep\_sorted} & \texttt{causal\_depth} & \texttt{reverse} & \multirow{3}{*}{$\rho$} \\
 & \footnotesize\itshape natural & \footnotesize\itshape dep.-respecting & \footnotesize\itshape time-major + & \footnotesize\itshape time-major + & \footnotesize\itshape full & \\
 & \footnotesize\itshape time-major & \footnotesize\itshape topological & \footnotesize\itshape dep.-based & \footnotesize\itshape depth-based & \footnotesize\itshape reversal & \\
\midrule
\textsc{ReLU}            & 73.3 & \textbf{100.0} & 84.6 & 63.0 & 0.0 & 0.90 \\
\textsc{Hidden-cause}    & 63.8 & \textbf{96.3}  & 83.8 & 40.0 & 0.5 & 0.90 \\
\textsc{Coupled ring}    & \textbf{99.8} & 99.5 & 99.8 & 99.9 & 32.6 & 0.72 \\
\textsc{Modular}         & 100.0 & 0.0 & \textbf{100.0} & 0.0 & 0.0 & 0.00 \\
\textsc{Sine}            & 98.6 & 99.9 & \textbf{100.0} & 99.2 & 0.0 & 0.70 \\
\textsc{Mixed nonlinear} & 4.8 & \textbf{1.4} & 64.5 & 10.7 & 0.0 & 0.00 \\
\bottomrule
\end{tabular}
\end{table}

Loss profiling replaces exhaustive full runs of a larger model with a single short run of a small model. \Cref{tab:delay-variants-summary} reports the success rates of generating complete sequences by models trained with each of the five candidates and the Spearman correlation between profile and retrain ranks.

\vspace{1mm}
\noindent\textbf{Results.} The learnability varies substantially even among topological sorts.
Loss profiling successfully discovers a learning-friendly order on five of the six tasks (indicated by bold entries in \cref{tab:delay-variants-summary}); only \textsc{Mixed nonlinear} fails, where early-stage losses are nearly tied (2.46 vs.\ 2.48;\ \cref{tab:delay-profile-losses}) and the rank-1 prediction (1.4\%) misses the actual best (64.5\%).
The Spearman correlation $\rho$ measures agreement between the profiling rank and the retrain rank over the five candidates; it is strong on four tasks ($\rho \geq 0.70$). On \textsc{Modular}, $\rho = 0$ comes from saturation-driven ties (many orders collapse to $0\%$ or $100\%$;\ all five profile losses lie within a $0.02$ band, see \cref{tab:delay-profile-losses}), not from true mismatch, and the rank-1 prediction is still a $100\%$ order. Results at larger $T$ on \textsc{ReLU} are reported in \cref{app:delay-original-scaling}.




\subsubsection{Beyond hand-designed candidates.}
The previous section showed that loss profiling can discover a learning-friendly order from five hand-designed topological sorts in \cref{tab:delay-variants-summary}. Next, we run a global-search experiment to search for useful orders beyond this fixed pool. We dropped the local stage as it is costly for the target length $36$. 

\vspace{1mm}
\noindent\textbf{Setup.}
The global stage uses two initializations: \texttt{topo} (100 random topological sorts, topology-aware) and \texttt{vars} (24 within-time orderings of $(a,b,c,d)$, topology-agnostic). The former assumes the dependency graph is known; the latter does not, only using the time-induced block initialization. The discovered order is retrained with the same 6-layer GPT-2 protocol as above.

\begin{wraptable}{r}{0.48\linewidth}
  \vspace{-1mm}
  \centering
  \small
  \setlength{\tabcolsep}{2pt}
  \caption{Global-search success rates (\%) on five delay networks ($T=11$). \textbf{Bold} marks the better of \texttt{topo}/\texttt{vars} per task.}
  \label{tab:delay-variants-global-search}
  \begin{tabular}{lccc}
  \toprule
  \textbf{Task} & \textbf{Hand-des.\ best} & \texttt{topo} & \texttt{vars} \\
  \midrule
  \textsc{Hidden-cause}    & 96.3 & \textbf{69.1} & 60.4 \\
  \textsc{Coupled ring}    & 99.9 & 99.1 & \textbf{99.8} \\
  \textsc{Modular}         & 100.0 & 0.0 & 0.0 \\
  \textsc{Sine}            & 100.0 & 99.3 & \textbf{100.0} \\
  \textsc{Mixed nonlinear} & 64.5 & \textbf{76.1} & 6.6 \\
  \bottomrule
  \end{tabular}
  \vspace{-5mm}
\end{wraptable}
\vspace{1mm}
\noindent\textbf{Results.}
\Cref{tab:delay-variants-global-search} compares discovered orders against the best hand-designed candidate. For \textsc{Coupled ring} and \textsc{Sine}, the discovered orders nearly match the best. Notably, topology-agnostic initialization (\texttt{vars}) performs on par with topology-aware one (\texttt{topo}) on these tasks and on \textsc{Hidden-cause}. For \textsc{Mixed nonlinear}, \texttt{topo} finds an order beyond the hand-designed best. \textsc{Modular} replicates the saturation issue from \cref{tab:delay-variants-summary} (see \cref{app:delay-global-search}).

\section{Conclusion}\label{sec:conclusion}
This study addressed reordering decoder input tokens for Transformers in sequential computation. The proposed method performs short-term training on candidate orders and selects those with the fastest loss drop; a two-stage scheme of global block-level exploration and local refinement handles the factorial search space.
On six order-sensitive tasks, it lifts success rates from about 10\,\% to near 100\,\%, and on \textsc{Prod} it rediscovers the reverse-digit order reported in prior work.
A case study on delay dynamical systems further shows transfer beyond single-variable recurrences, where the global search even surpasses the best hand-designed candidate.

\vspace{1mm}
\noindent\textbf{Limitations and future work.} 
The local stage of the proposed method is not cheap for long sequences. Although the global stage still finds effective orders, improving the local stage is future work. We suspect the global-local search can be further improved by exploiting the design of sorting algorithms. Extending this to general inputs, such as natural languages, is another future direction.


\section*{Acknowledgments}
This research was partially supported by JST PRESTO Grant Number JPMJPR24K, JST BOOST Program Grant Number JPMJBY24C6, and JSPS Program for Forming Japan’s Peak Research Universities (J-PEAKS) Grant Number JPJS00420230002, Mitsubishi Electric Information Technology R\&D Center, and JSPS KAKENHI Grant Number 26K02996.

\newpage
\bibliographystyle{plainnat}
\bibliography{refs}

\clearpage
\appendix

\newpage
\appendix
\onecolumn

\section{Visualizing the global--local pipeline}\label{app:pipeline-vis}
\begin{figure}[h]
  \centering
  \includegraphics[width=1.0\linewidth]{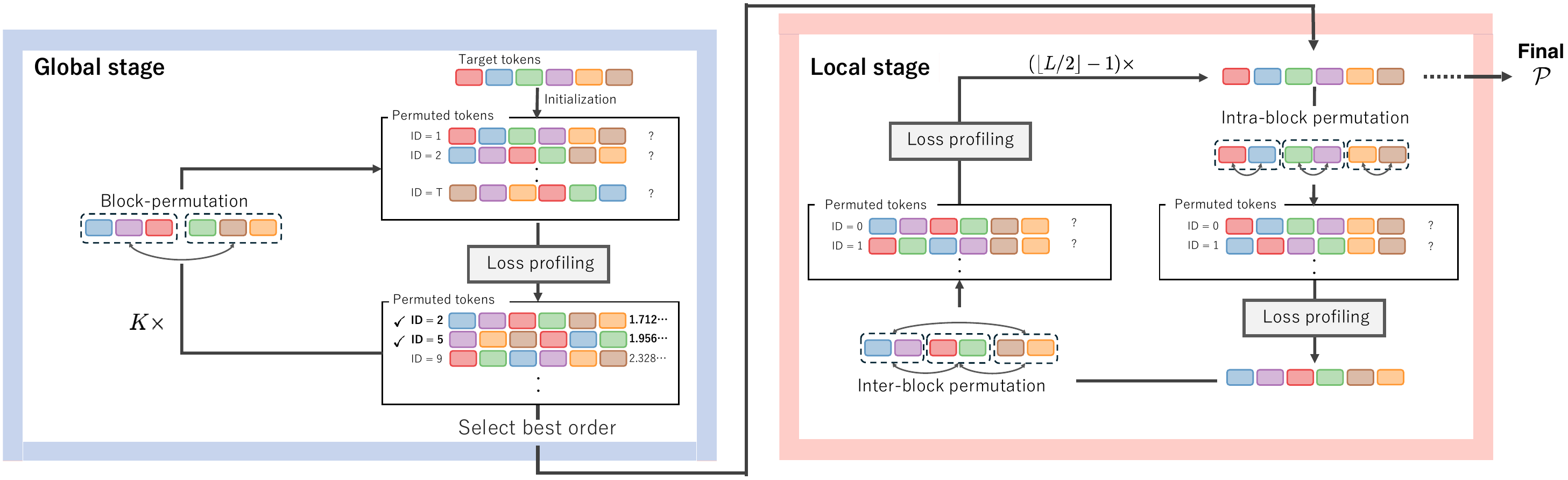}
  \caption{Overview of the exploration. \textbf{Global stage} (left panel) generates $T$ candidate permutations by swapping the sequence at the macro-level, exchanging token blocks to quickly spot coarse, learning-friendly orders. \textbf{Local stage} (right panel) further permutes the tokens for each target block, refining the sequence to discover a final permutation that maximizes learning ease.}
  \label{fig:HPS_flow}
\end{figure}

\Cref{fig:HPS_flow} illustrates the hierarchical exploration procedure described in \cref{sec:proposed-method}.

\section{Visualizing attention map}\label{app:vis_attention_map}
\begin{figure}[h]
  \centering

  \begin{subfigure}{0.45\textwidth}
    \includegraphics[width=\linewidth]{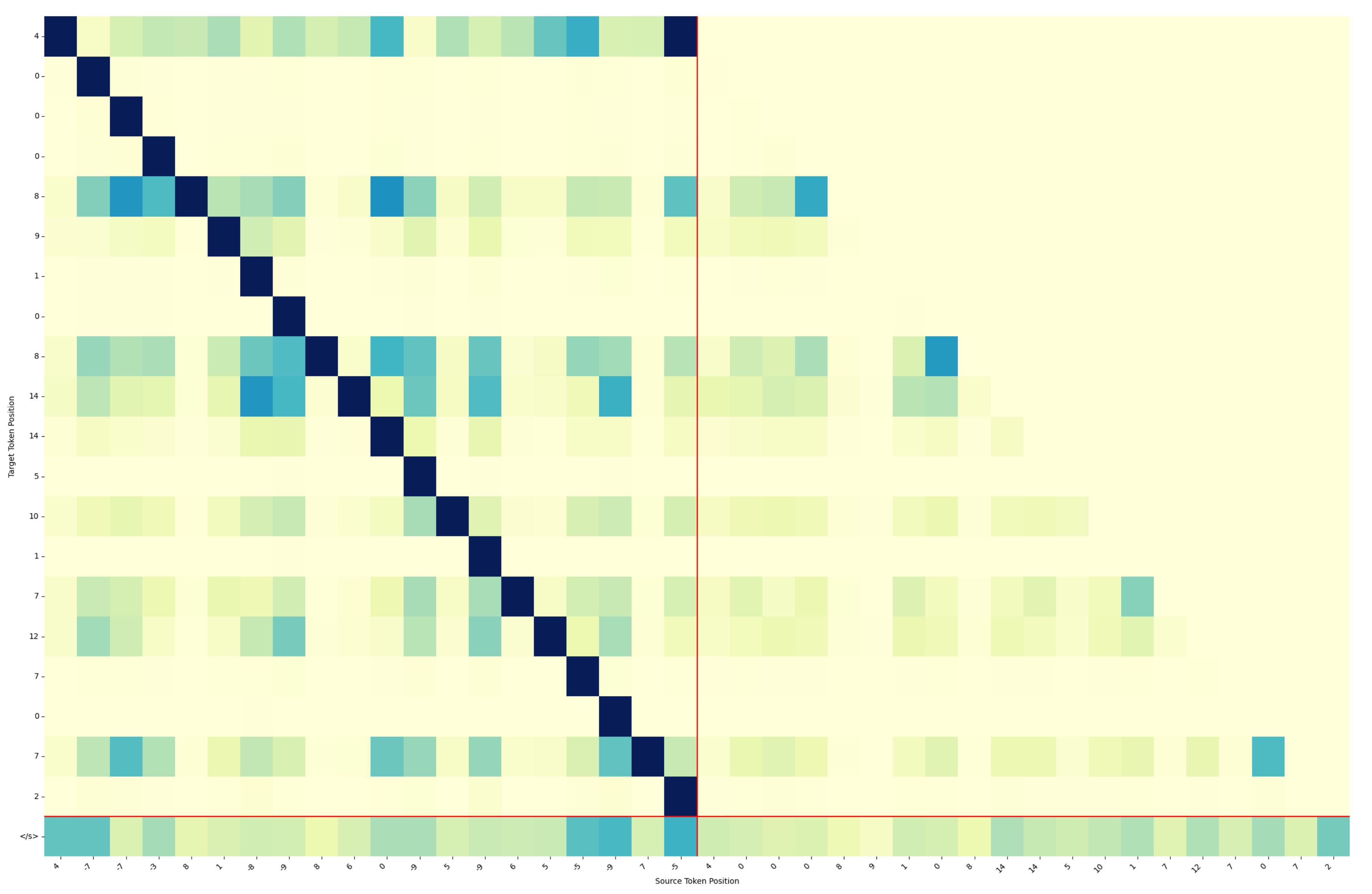}
    \caption{Forward order}
  \end{subfigure}
  \hfill
  \begin{subfigure}{0.45\textwidth}
    \includegraphics[width=\linewidth]{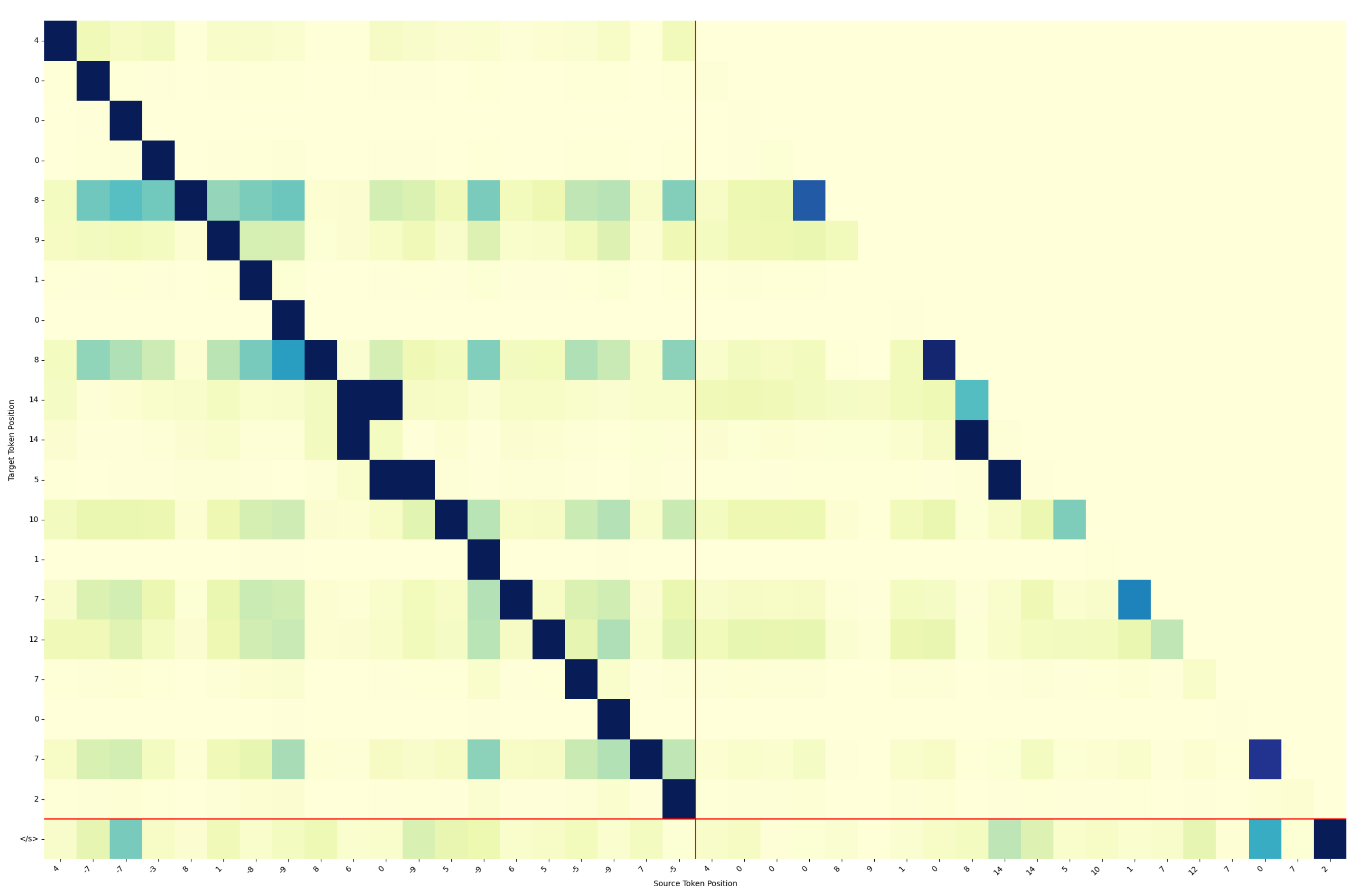}
    \caption{one permuted order}
  \end{subfigure}

  \vskip\baselineskip

  \begin{subfigure}{0.45\textwidth}
    \includegraphics[width=\linewidth]{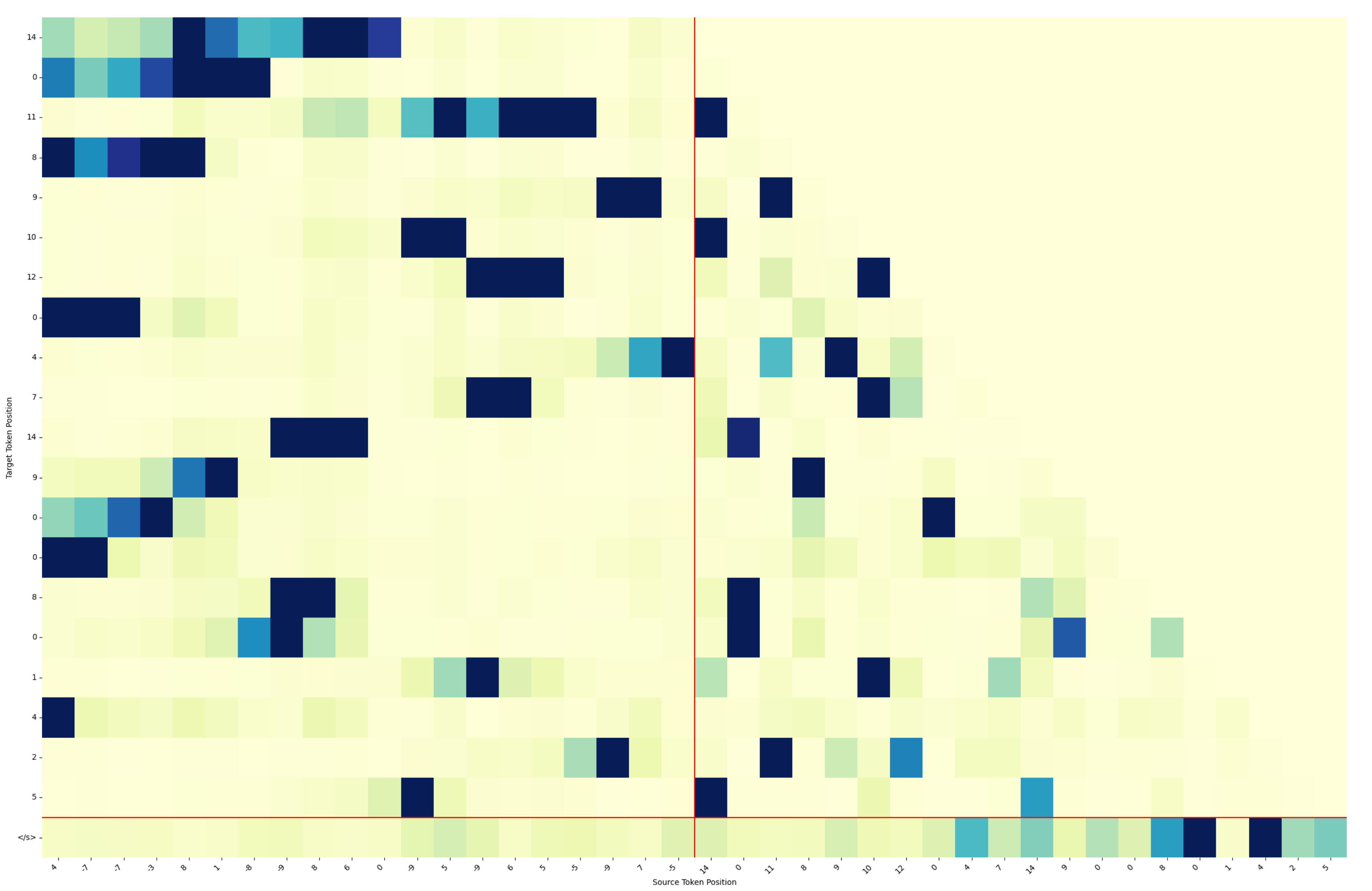}
    \caption{Random order}
  \end{subfigure}
  \hfill
  \begin{subfigure}{0.45\textwidth}
    \includegraphics[width=\linewidth]{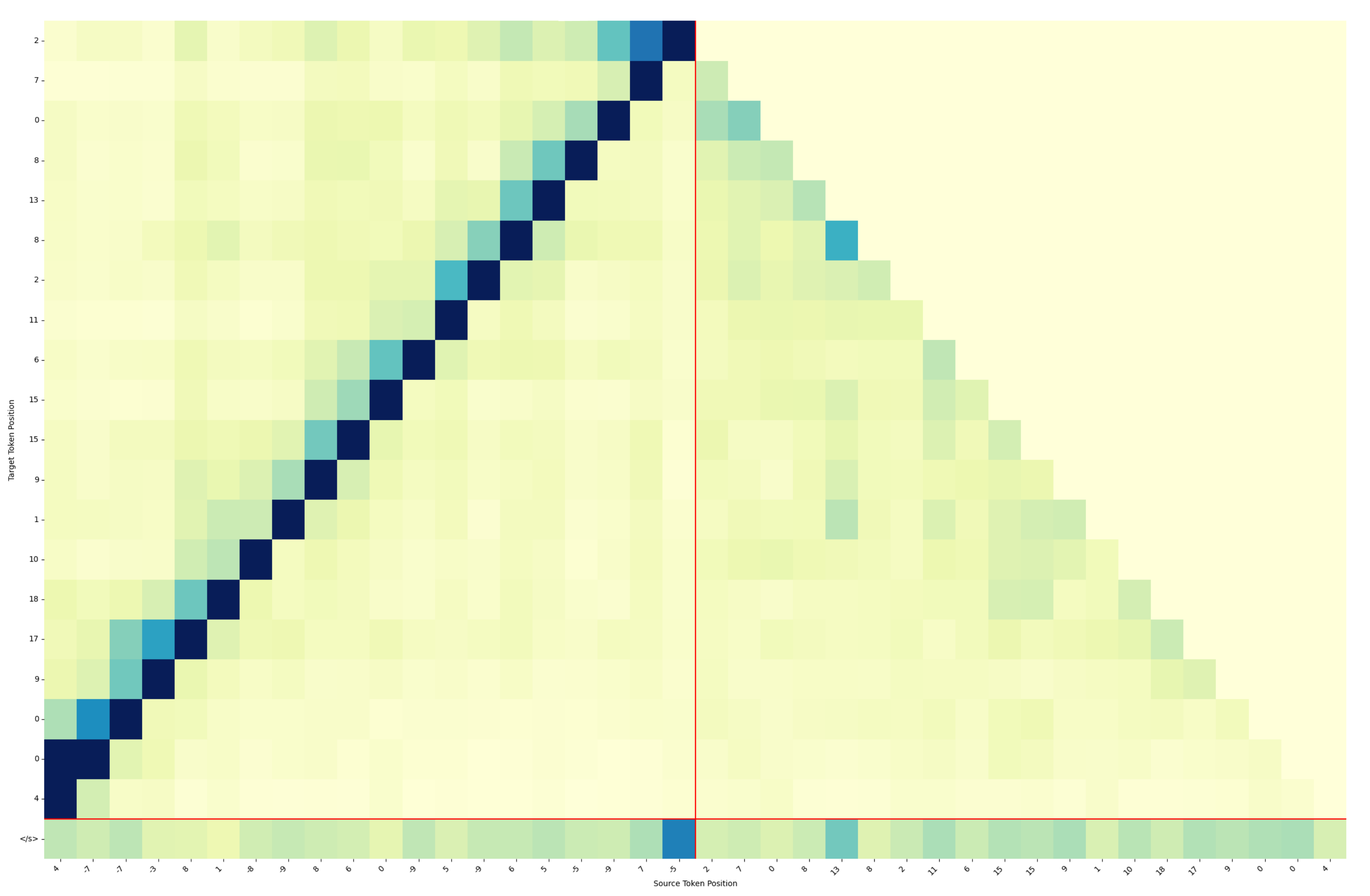}
    \caption{Reverse order}
  \end{subfigure}

  \caption{
    Attention matrices from models trained with four different target orders in the \textsc{ReLU} task.
  }
  \label{fig:relu_attention_four_orders}
\end{figure}

\begin{figure}[h]
  \centering
  \includegraphics[width=\textwidth]{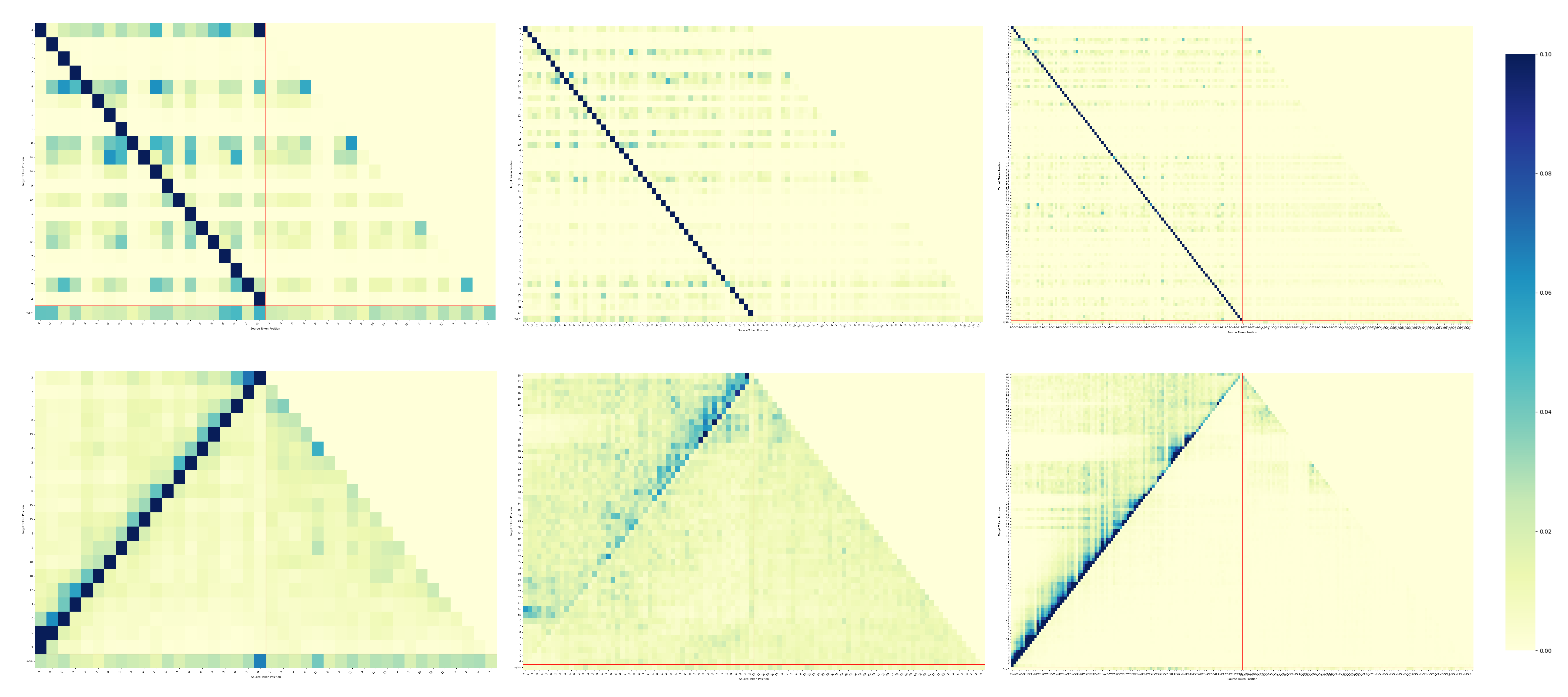}
  \caption{
    Differences in the attention matrices for the \textsc{ReLU} task between forward and reverse orderings. 
    The top three matrices correspond to models trained with forward order, and the bottom three with reverse order. 
    Each pair of matrices shows results for input lengths $n = 20$, $50$, and $100$, respectively.
  }
  \label{fig:relu_attention_forward_reverse}
\end{figure}

We present the attention maps obtained when training a Transformer on our proposed \textsc{ReLU} task with datasets reordered in different ways.
For this analysis, we use a GPT-2 model with a single layer and a single attention head.
\Cref{fig:relu_attention_four_orders} shows the attention maps for target length $L=20$ under four target orders.
The forward and reverse orders are defined in~\cref{sec:order-sensitive-tasks}.
The one-permuted order swaps exactly one pair of adjacent target tokens, whereas the random order is a random permutation of the forward sequence.
\Cref{fig:relu_attention_forward_reverse} illustrates how the attention maps change as the target length increases.

\begin{table}[h]
\centering
\caption{Attention sparsity $S$ across target orders. A smaller value of $S$ indicates greater sparsity.}
\label{tab:attn_entropy}
\begin{tabular}{llcc}
\toprule
\textbf{Task} & \textbf{Target length} &
\multicolumn{2}{c}{\textbf{Sparsity}}\\
\cmidrule(lr){3-4}
& & \textbf{Forward} & \textbf{Reverse}\\
\midrule
\multirow{3}{*}{\textsc{ReLU}}
  & $L=20$ & \textbf{1.160} & 1.640 \\
  & $L=50$ & \textbf{1.462} & 4.319 \\
  & $L=100$& \textbf{1.687} & 3.195 \\
\midrule
\multirow{3}{*}{\textsc{Square}}
  & $L=20$ & \textbf{1.117} & 1.531 \\
  & $L=50$ & \textbf{1.773} & 1.914 \\
  & $L=100$ & \textbf{1.407} & 1.990 \\
\midrule
\multirow{3}{*}{\textsc{Index}}
  & $L=13, d=2$ & \textbf{0.848} & 2.574 \\
  & $L=13, d=4$ & \textbf{0.887} & 1.486 \\
  & $L=13, d=8$ & \textbf{1.116} & 1.596 \\
\bottomrule
\end{tabular}
\end{table}

\section{Soft-permutation optimization via attention sparsity} \label{sec:attn_based_optimization}
\subsection{Analysis of information leakage in naive soft-permutation optimization}
\label{app:soft_perm_leakage}

We address the fundamental limitation of optimizing a soft permutation matrix $\tilde{P} \in [0, 1]^{L \times L}$ jointly with the model parameters.
We formulate the naive optimization objective as the minimization of the empirical risk:
\begin{align}
    \min_{\theta, \tilde{P}} \ \frac{1}{m} \sum_{i=1}^m \ell(\cT_{\theta}, X_i, Y_i \tilde{P}).
\end{align}

\begin{figure}[t]
    \centering
    \includegraphics[width=0.75\linewidth]{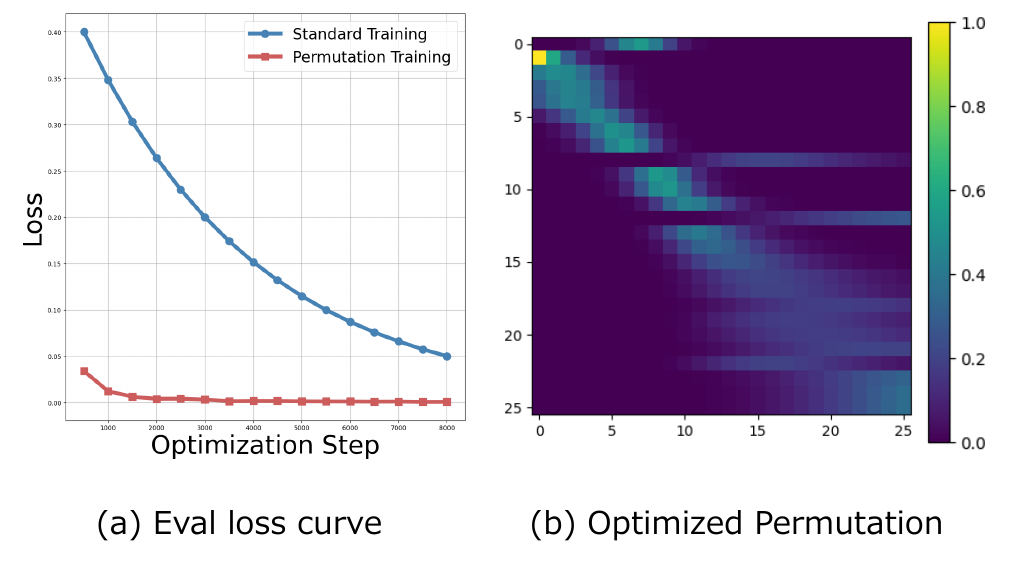}
    \caption{Failure mode of naive soft-permutation optimization. (a) Training loss drops immediately due to leakage. (b) The learned soft permutation has off-diagonal weights near the main diagonal, indicating access to future tokens.}
    \label{fig:leak-loss-and-perm}
\end{figure}

However, as shown in \cref{fig:leak-loss-and-perm}, such an approach leads to an immediate loss drop at the early stage of training. This creates a trivial solution where the soft permutation $\tilde{P}$ causes information leakage from future tokens. Specifically, each token in the reordered target sequence $Y_i \tilde{P}$ becomes a soft mixture of all the tokens in the original sequence $Y_i$.
In an auto-regressive setting, this mixture allows the model to access the ground truth of the next token (which is contained in the future positions of $Y_i$) through the non-zero off-diagonal entries of $\tilde{P}$, thereby undermining the next-token prediction task.

\Cref{fig:leak-loss-and-perm} (b) visualizes the learned soft permutation matrix obtained from this naive training. The matrix exhibits sparse off-diagonal weights clustered around the main diagonal. These weights indicate that the optimization process finds shortcuts to ``peek'' at the immediate future tokens rather than discovering a meaningful global causal ordering. This phenomenon motivates the need for alternative approaches, such as the attention sparsity regularization or the proposed discrete search pipeline, to prevent such leakage.


\subsection{Analysis of attention sparsity.}
\label{app:analysis_attn_sparsity}

To address the challenges of permutation optimization, we undertake a more detailed analysis of the attention mechanism.
Intuitively, when the target order is learning-unfriendly, the local causal structure of the sequence is broken, meaning that more input and output tokens become relevant to predicting the next token.
Conversely, for a learning-friendly order, we expect the attention map to be \emph{sparser}, as the model can rely on specific local dependencies.

Let the query and key matrices be $Q, K \in \mathbb{R}^{L^{\prime} \times d_{\mathrm{emb}}}$, where $L^{\prime}$ is the decoder-input length and $d_{\mathrm{emb}}$ is the embedding dimension.
The self-attention weights are given by
\begin{align}
    A = \mathrm{Softmax} \left( \frac{QK^{\top}}{\sqrt{d_{\mathrm{emb}}}} \right) \in \mathbb{R}^{L^{\prime} \times L^{\prime}},
\end{align}
where $\mathrm{Softmax}(\cdot)$ is applied row-wise.
Because each row of $A = (a_{ij})_{ij}$ is a probability vector, we define the mean sparsity $S$ using the Shannon entropy:
\begin{align}
    S = -\frac{1}{L^{\prime}} \sum_{i=1}^{L^{\prime}} \sum_{j=1}^{L^{\prime}} a_{ij} \log a_{ij}.
\end{align}
We compute $S$ for models trained on both the forward (learning-friendly) and reverse (learning-unfriendly) orders of the order-sensitive tasks (\cref{sec:order-sensitive-tasks}).
\Cref{tab:attn_entropy} shows that the forward order consistently yields lower $S$. Since a smaller $S$ directly corresponds to lower entropy (i.e., higher sparsity), this confirms that learning-friendly orders produce sparser attention maps.
Representative heat maps are provided in \cref{app:vis_attention_map}.

Because $S$ is derived solely from the learned attention weights, it is independent of the language-model loss and can serve as an orthogonal diagnostic metric.
We explored optimizing permutations using an additional sparsity regularizer that explicitly rewards low-entropy attention.
However, even with this bias, the optimizer failed to discover the learning-friendly order and instead converged to interleaved permutations, suggesting that sparsity guidance alone is insufficient to solve the permutation search in difficult regimes.

\begin{figure}[t]
  \centering
  \includegraphics[width=0.8\linewidth]{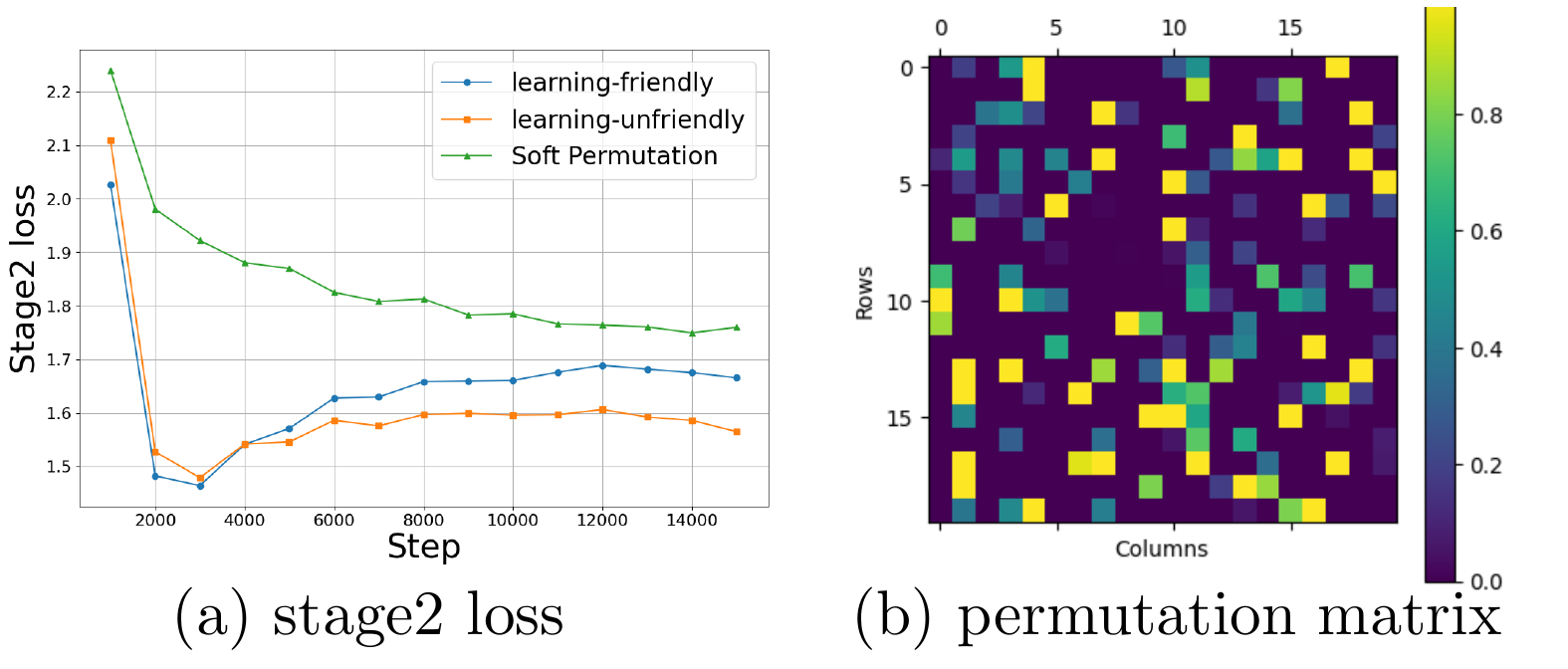}
  \caption{(a) Comparison of Stage 2 loss under fixed learning‐friendly order, fixed learning‐unfriendly order, and learned soft permutation. (b) shows a visualization of the learned soft permutation.}
  \label{fig:attention-entropy-results}
\end{figure}

Specifically, we formulated a soft-permutation optimization method based on attention sparsity.
In our two-stage strategy, we first optimize the Transformer parameters $\theta$ by minimizing the standard sequence-modeling loss over the training set:
\begin{equation}\label{eq:stage1}
    \min_{\theta}\ \frac{1}{m}\sum_{i=1}^{m}\ell \bigl(\mathcal{T}_{\theta},\,X_{i},\,Y_{i}\bigr).
\end{equation}
Next, denoting by $A(\tilde{P}) = [a_{ij}]$ the attention map produced when the target sequence is fed as $Y_{i}\tilde{P}$ into the Transformer, we optimize the soft permutation $\tilde{P}$ by minimizing the total attention entropy:
\begin{equation}\label{eq:stage2}
    \min_{\tilde{P}}\ S(A(\tilde{P})) = \min_{\tilde{P}} \left( -\frac{1}{L^{\prime}} \sum_{i=1}^{L^{\prime}} \sum_{j=1}^{L^{\prime}} a_{ij} \log a_{ij} \right).
\end{equation}
In our experiments, we alternated between these two optimization stages.
\Cref{fig:attention-entropy-results} compares the Stage 2 loss (entropy) under three conditions: the fixed learning-friendly order, the fixed learning-unfriendly (reverse) order, and the learned soft permutation.
We observe that the soft permutation optimization does not significantly reduce the entropy-based loss relative to the fixed orders, nor does it converge to a genuinely structured ordering. 
This indicates that attention sparsity, while a useful diagnostic metric, is difficult to use as a direct, reliable objective for permutation optimization due to optimization landscapes or trivial solutions.

\section{Permutation search via evolutionary strategy} \label{app:permutation_search_ES}
This section summarizes the evolutionary strategy~(ES) baseline that we ran in parallel with our proposed method to search the permutation space.
Each individual is a permutation $P$; its fitness is the (negative) early-stage training loss of a Transformer trained with that order, so that permutations that are easier to learn receive higher scores.
The ES is controlled by the population size \(N_p\), crossover probability \(N_c\), mutation probability \(N_m\), number of generations \(N_g\), tournament size \(N_t\), and elitism ratio \(N_r\), and proceeds as follows:
\begin{enumerate}[leftmargin=1.5em,label=(\arabic*)]
    \item \textbf{Population initialization:} sample \(N_p\) random permutations.
    \item \textbf{Selection:} pick parents via tournament selection with size~\(N_t\).
    \item \textbf{Crossover:} with probability~\(N_c\), apply partially–mapped crossover to each selected pair.
    \item \textbf{Mutation:} with probability~\(N_m\), swap two positions in the offspring permutation.
    \item \textbf{Elitism:} evaluate every individual by
          \[
              \mathrm{fitness}(P)
              = 
              -\frac{1}{m'}\sum_{i=1}^{m'}~
              \ell\!\bigl(\mathcal{T}_{\theta},\, X_i,\, Y_i P\bigr),
          \]
          and copy the top \(N_r\) fraction to the next generation.
    \item \textbf{Termination:} stop when \(N_g\) generations have been processed.
\end{enumerate}

\Cref{tab:es_success} lists the permutation identified by the evolution strategy (ES) and the performance obtained when the model is retrained using that order.


\begin{table}[t]
  \small
  \centering
  \caption{Success rate obtained when the Transformer is retrained on the permutations discovered by the ES.}
  \label{tab:es_success}
  \begin{tabular}{lllcc}
    \toprule
    \textbf{Task} & \textbf{Input length} & \textbf{ES-discovered order} &
    \multicolumn{2}{c}{\textbf{Success rate (\%)}}\\
    \cmidrule(lr){4-5}
    & & & \textbf{Retrain} & \textbf{Reverse} \\
    \midrule
    \multirow{3}{*}{\textsc{ReLU}}
      & $L=5$  & [2,\,1,\,0,\,4,\,3]                                                     & 26.9 & 10.4 \\
      & $L=10$ & [0,\,1,\,2,\,3,\,4,\,5,\,6,\,7,\,8,\,9]                                 & 100  &  3.5 \\
      & $L=20$ & [6,\,7,\,9,\,8,\,12,\,11,\,13,\,18,\,17,\,14,\,16,\,15,\,19,\,5,\,10,\,1,\,0,\,3,\,4,\,2] &  9.2 & 0.7 \\
    \midrule
    \multirow{3}{*}{\textsc{Square}}
      & $L=5$  & [1,\,2,\,3,\,4,\,0]                                                     & 100  & 21.5 \\
      & $L=10$ & [3,\,4,\,5,\,6,\,7,\,8,\,9,\,1,\,0,\,2]                                 & 99.9 & 6.5 \\
      & $L=20$ & [9,\,10,\,11,\,12,\,13,\,14,\,2,\,3,\,4,\,5,\,6,\,15,\,16,\,17,\,18,\,19,\,0,\,1,\,7,\,8] &  5.2 &  0.1 \\
    \bottomrule
  \end{tabular}
\end{table}

\section{Details on order-sensitive tasks}
\label{app:order_sensitive_proofs}
\subsection{Order-sensitive tasks}
\label{app:order-sensitive-tasks}

Let $\Sigma$ be the set of tokens. As defined in \cref{sec:unraveling-CoT}, we consider a task to be a deterministic mapping $F: \Sigma^\ast \to \Sigma^L$. For an input sequence $X \in \Sigma^\ast$, the target output is $Y = F(X) = [y_1, \dots, y_L] \in \Sigma^L$.
We denote the set of all permutations over indices $\{1, \dots, L\}$ by $S_L$. For a fixed permutation $\pi \in S_L$, the reordered target sequence is denoted by $Y^{(\pi)} := [y_{\pi(1)}, \dots, y_{\pi(L)}]$.

To rigorously quantify the computational complexity of generating $Y$ under a specific order, we first establish a cost model. Let $\Omega$ be a fixed set of primitive operators (e.g., $\{+, \times, \mathrm{mod}, \mathrm{ReLU}, \dots\}$), where each operator incurs a unit cost.

\vspace{1mm}
\noindent\textbf{Autoregressive Context Constraint.}
We define a sequential generation algorithm respecting order $\pi$ as a tuple $A_\pi = (\phi_1, \dots, \phi_L)$, where each $\phi_t$ produces the next token $y_{\pi(t)}$ from the information available at autoregressive decoding step $t$.
Specifically, for every $t \in \{1, \dots, L\}$, the output must be computed from the input $X$ and the previously generated tokens $Y_{<t}^{(\pi)} := [y_{\pi(1)}, \dots, y_{\pi(t-1)}]$:
\begin{align}
    y_{\pi(t)} = \phi_t\left(X, Y_{<t}^{(\pi)}\right).
\end{align}
This constraint is about the autoregressive interface, not about the absence of internal representations in a Transformer.
KV caches may store hidden representations of the prefix for efficient decoding, but they do not add independent writable memory or allow arbitrary intermediate computation results from earlier prediction steps to be passed forward as an additional scratchpad.
Thus, any information that can influence later predictions must be encoded in the original input or in the tokens generated so far.

The operation count of such an algorithm, denoted by $\mathrm{Ops}(A_\pi; X)$, is the total number of operators from $\Omega$ invoked across all functions $\phi_1, \dots, \phi_L$.
The minimal sequential cost is defined as:
\begin{align}
    \mathcal{C}_\pi(X) := \min_{A_\pi} \ \mathrm{Ops}(A_\pi; X),
\end{align}
where the minimization is over all algorithms $A_\pi$ satisfying the autoregressive context constraint.


\vspace{1mm}
\noindent\textbf{Order-Sensitivity.}
We characterize the difficulty of a task based on the disparity in computational cost between different generation orders.

\begin{definition}[Order sensitivity]
\label{def:order-sensitive}
Let $\mathcal{D}$ be a distribution over inputs $X$. We define the \emph{order-sensitivity ratio} $\rho(L)$ for a task $F$ as the ratio of expected minimal costs between the worst-case order $\sigma$ and the best-case order $\pi$:
\begin{align}
    \rho(L) := \sup_{\pi, \sigma \in S_L} \frac{\mathbb{E}_{X \sim \mathcal{D}}[\mathcal{C}_\sigma(X)]}{\mathbb{E}_{X \sim \mathcal{D}}[\mathcal{C}_\pi(X)]}.
\label{eq:order-sensitive-ratio}
\end{align}
We classify the task $F$ as \textbf{order-sensitive} if this ratio grows asymptotically with the sequence length $L$, i.e.,
\begin{align}
    \lim_{L \to \infty} \rho(L) = \infty.
\end{align}
\end{definition}


\subsection{Proof of order-sensitivity}
\label{app:proof-order-sensitivity}

We verify that the six tasks (\textsc{ReLU}, \textsc{Square}, \textsc{MLP}, \textsc{Sine}, \textsc{Cubic}, \textsc{Triangle}) proposed in \cref{sec:experiments} satisfy the definition of order-sensitivity. Specifically, we demonstrate that they are $\rho$-order-sensitive with $\rho(L) = \Theta(L)$.

\begin{proposition}
\upshape
The tasks \textsc{ReLU}, \textsc{Square}, \textsc{MLP}, \textsc{Sine}, \textsc{Cubic}, and \textsc{Triangle} are order-sensitive.
\end{proposition}

\begin{proof}
Let $\pi_{\mathrm{fwd}}$ be the forward order (identity permutation) and $\pi_{\mathrm{rev}}$ be the reverse order. We analyze the minimal sequential costs under the autoregressive context constraint defined in \cref{app:order-sensitive-tasks}.

\vspace{1mm}
\noindent\textbf{Forward Complexity.}
In the forward order, the sequence is generated via a deterministic recurrence $y_1 = x_1$ and $y_i = f(x_i, y_{i-1})$ for $i=2, \dots, L$.
The optimal autoregressive algorithm under this constraint simply executes this recurrence step-by-step. Since $y_{i-1}$ is available in the immediate context $Y_{<i}^{(\pi_{\mathrm{fwd}})}$, the cost to compute $y_i$ is constant, denoted by $c_f$. Thus,
\begin{align}
    \mathcal{C}_{\pi_{\mathrm{fwd}}}(X) = \sum_{i=1}^L c_f = \Theta(L).
\end{align}

\vspace{1mm}
\noindent\textbf{Reverse Complexity and Verification Cost.}
In the reverse order, the algorithm must generate $y_L, y_{L-1}, \dots, y_1$. At step $t$ (where the target is $y_{L-t}$), the available information is the input $X$ and the suffix $y_L, \dots, y_{L-t+1}$. The algorithm must determine the preceding token $y_{L-t-1}$.

However, for all proposed tasks, the forward map $f(\cdot, y_{i-1})$ is non-injective with respect to $y_{i-1}$. Let $g_{x_i}(z) := f(x_i, z)$. The inverse map $g_{x_i}^{-1}(y_i)$ produces a set of candidates $\mathcal{S}_i = \{z \mid f(x_i, z) = y_i\}$ rather than a single value.
For example, in \textsc{Square} ($y_i = x_i^2 + y_{i-1}^2 \bmod N$), $|\mathcal{S}_i| \ge 2$ due to quadratic residues. Similarly, \textsc{ReLU} and \textsc{Triangle} map multiple inputs to the same output.

To identify the correct predecessor $y^*_{i-1} \in \mathcal{S}_i$ that is consistent with the initial condition $y_1=x_1$, the algorithm must verify which candidate lies on the valid trajectory originating from $x_1$.
Under the autoregressive context constraint, the function $\phi_t$ cannot access intermediate computation results from previous steps unless they have been emitted as tokens. Therefore, the only way to deterministically verify a candidate is to recompute the forward trajectory from $x_1$ up to $i-1$ and check if the result matches the candidate.
This implies that generating the token at position $i$ effectively incurs a cost of forward simulation from the start, which is proportional to $i$.

The minimal sequential cost for the reverse order is thus bounded by the sum of these verification costs:
\begin{align}
    \mathcal{C}_{\pi_{\mathrm{rev}}}(X) \approx \sum_{i=1}^{L} i = \frac{L(L+1)}{2} = \Theta(L^2).
\end{align}

\vspace{1mm}
\noindent\textbf{Ratio.}
Combining the above, the order-sensitivity ratio is:
\begin{align}
    \frac{\mathbb{E}[\mathcal{C}_{\pi_{\mathrm{rev}}}(X)]}{\mathbb{E}[\mathcal{C}_{\pi_{\mathrm{fwd}}}(X)]} 
    \approx \frac{\Theta(L^2)}{\Theta(L)} 
    = \Theta(L).
\end{align}
This satisfies the condition $\lim_{L \to \infty} \rho(L) = \infty$, confirming the tasks are order-sensitive.
\end{proof}

\section{Details on experimental settings}
\label{app:experimental_settings_details}

In this section, we provide concrete examples of the datasets introduced in \cref{sec:training_setup} and visualize the permutation sets used in our experiments.

\subsection{Dataset examples}\label{app:dataset_examples}
We present representative samples for the seven order-sensitive tasks described in \cref{sec:order-sensitive-tasks}.
\Cref{tab:dataset_samples} summarizes the correspondence between the input $X$ and the target $Y$, where $Y$ represents the sequence in the forward (learning-friendly) order.
Note that for the \textsc{Prod} task, the input consists of two integers $a$ and $b$. For the \textsc{MLP} task, we use
\begin{align}
\operatorname{MLP}(u,v)=W_2\,\operatorname{ReLU}\!\left(W_1 [u,v]^\top + b_1\right)+b_2,
\end{align}
where $W_1\in\mathbb{R}^{h\times 2}$, $b_1\in\mathbb{R}^{h}$, $W_2\in\mathbb{R}^{1\times h}$, and $b_2\in\mathbb{R}$ are sampled once (He initialization;~\citep{he2015delving}) and then fixed throughout dataset generation.

\begin{table*}[h]
\small
\centering
\caption{Representative input–output samples for each task}
\begin{tabular}{c|l|l}
\toprule
\textbf{Task} & \textbf{Input} & \textbf{Target} \\
\midrule
\multirow{5}{*}{\textsc{ReLU}, $L=50$} 
  & $X = (4,\ -7,\ -7,\ -3,\ 8,\ 1,\ -8,\ -9,\ 8,\ 6$ & $Y = (4,\ 0,\ 0,\ 0,\ 8,\ 9,\ 1,\ 0,\ 8,\ 14$ \\
  & $0,\ -9,\ 5,\ -9,\ 6,\ 5,\ -5,\ -9,\ 7,\ -5$ & $14,\ 5,\ 10,\ 1,\ 7,\ 12,\ 7,\ 0,\ 7,\ 2$ \\
  & $8,\ -6,\ -7,\ -2,\ -7,\ 6,\ 7,\ -2,\ 0,\ -6$ & $10,\ 4,\ 0,\ 0,\ 0,\ 6,\ 13,\ 11,\ 11,\ 5$ \\
  & $-3,\ -8,\ -7,\ -8,\ 3,\ -1,\ -6,\ 1,\ -4,\ -9$ & $2,\ 0,\ 0,\ 0,\ 3,\ 2,\ 0,\ 1,\ 0,\ 0$ \\
  & $2,\ -7,\ 1,\ 4,\ 9,\ -5,\ 6,\ 2,\ 3,\ -3)$ & $2,\ 0,\ 1,\ 5,\ 14,\ 9,\ 15,\ 17,\ 20,\ 17)$ \\

\midrule
\multirow{5}{*}{\textsc{Square}, $L=50$} 
  & $X = (-5,\ -9,\ 8,\ 7,\ 8,\ -7,\ 5,\ -9,\ -6,\ 9$ & $Y = (-5,\ 2,\ 2,\ 6,\ -4,\ -1,\ -2,\ 0,\ 8,\ 3$ \\
  & $-2,\ -8,\ 6,\ -7,\ 2,\ -7,\ -6,\ -5,\ -5,\ 7$ & $4,\ -5,\ -5,\ 8,\ 2,\ 6,\ 6,\ -5,\ 3,\ -8$ \\
  & $3,\ 6,\ -9,\ 1,\ 7,\ 0,\ -7,\ 7,\ -5,\ 0$ & $7,\ 0,\ -4,\ 8,\ 9,\ -4,\ -1,\ 3,\ 6,\ 8$ \\
  & $-2,\ 6,\ -1,\ -9,\ -6,\ -7,\ 0,\ 2,\ 7,\ -1$ & $2,\ -7,\ 3,\ 5,\ -5,\ 8,\ -2,\ -1,\ 3,\ 1$ \\
  & $1,\ -2,\ -6,\ -7,\ 5,\ 1,\ 9,\ -6,\ -3,\ -3 )$ & $-7,\ 6,\ 6,\ 0,\ -3,\ 1,\ -3,\ -2,\ 4,\ -3 )$ \\

\midrule
\multirow{5}{*}{\textsc{Triangle}, $L=50$} 
  & $X = (1,\ 17,\ 11,\ 18,\ 18,\ 5,\ 10,\ 12,\ 20,\ 11$ & $Y = (1,\ 2,\ 7,\ 5,\ 3,\ 12,\ 2,\ 6,\ 6,\ 3$ \\
  & $9,\ 12,\ 6,\ 11,\ 3,\ 13,\ 20,\ 12,\ 2,\ 16$ & $8,\ 0,\ 14,\ 5,\ 12,\ 5,\ 5,\ 3,\ 15,\ 11$ \\
  & $11,\ 12,\ 12,\ 9,\ 0,\ 11,\ 19,\ 3,\ 14,\ 7$ & $2,\ 6,\ 2,\ 9,\ 11,\ 2,\ 1,\ 16,\ 10,\ 3$ \\
  & $5,\ 10,\ 19,\ 14,\ 19,\ 8,\ 6,\ 17,\ 2,\ 20$ & $12,\ 2,\ 1,\ 5,\ 4,\ 8,\ 6,\ 3,\ 15,\ 15$ \\
  & $16,\ 19,\ 18,\ 1,\ 2,\ 13,\ 20,\ 19,\ 1,\ 0)$ & $11,\ 10,\ 8,\ 11,\ 7,\ 0,\ 0,\ 1,\ 18,\ 2)$ \\
\midrule
\multirow{5}{*}{\textsc{MLP}, $L=50$} 
  & $X = (1,\ 0,\ 16,\ 12,\ 12,\ 10,\ 2,\ 30,\ 24,\ 7$ & $Y = (1,\ 30,\ 16,\ 21,\ 19,\ 21,\ 0,\ 21,\ 14,\ 24$ \\
  & $22,\ 10,\ 9,\ 9,\ 7,\ 3,\ 14,\ 1,\ 1,\ 27$ & $14,\ 22,\ 23,\ 23,\ 26,\ 0,\ 26,\ 2,\ 29,\ 11$ \\
  & $6,\ 8,\ 19,\ 14,\ 27,\ 10,\ 10,\ 13,\ 21,\ 29$ & $25,\ 25,\ 15,\ 20,\ 13,\ 22,\ 22,\ 19,\ 16,\ 12$ \\
  & $9,\ 11,\ 23,\ 13,\ 8,\ 9,\ 4,\ 9,\ 3,\ 2$ & $23,\ 20,\ 15,\ 21,\ 24,\ 23,\ 30,\ 24,\ 0,\ 30$ \\
  & $27,\ 26,\ 25,\ 30,\ 24,\ 28,\ 3,\ 28,\ 17,\ 7)$ & $11,\ 17,\ 14,\ 13,\ 15,\ 13,\ 29,\ 11,\ 20,\ 25)$ \\
\midrule
\multirow{5}{*}{\textsc{Sine}, $L=50$} 
  & $X = (9,\ 1,\ 28,\ 31,\ 28,\ 27,\ 4,\ 20,\ 13,\ 25$ & $Y = (9,\ 25,\ 28,\ 3,\ 25,\ 4,\ 0,\ 4,\ 6,\ 25$ \\
  & $21,\ 5,\ 27,\ 10,\ 22,\ 28,\ 4,\ 15,\ 5,\ 1$ & $25,\ 25,\ 4,\ 25,\ 25,\ 28,\ 31,\ 25,\ 25,\ 25$ \\
  & $5,\ 2,\ 29,\ 29,\ 11,\ 9,\ 7,\ 17,\ 7,\ 30$ & $25,\ 3,\ 31,\ 28,\ 6,\ 25,\ 31,\ 31,\ 6,\ 4$ \\
  & $11,\ 24,\ 24,\ 20,\ 5,\ 1,\ 13,\ 21,\ 10,\ 27$ & $25,\ 6,\ 25,\ 3,\ 0,\ 6,\ 28,\ 6,\ 31,\ 25$ \\
  & $17,\ 3,\ 27,\ 16,\ 22,\ 31,\ 11,\ 15,\ 28,\ 3)$ & $25,\ 28,\ 6,\ 6,\ 28,\ 3,\ 25,\ 0,\ 28,\ 25)$ \\
\midrule
\multirow{5}{*}{\textsc{Cubic}, $L=50$} 
  & $X = (12,\ 10,\ 18,\ 15,\ 15,\ 7,\ 6,\ 5,\ 8,\ 18$ & $Y = (12,\ 9,\ 6,\ 3,\ 4,\ 14,\ 14,\ 13,\ 1,\ 0$ \\
  & $13,\ 3,\ 12,\ 11,\ 0,\ 18,\ 18,\ 13,\ 3,\ 5$ & $13,\ 15,\ 5,\ 3,\ 8,\ 17,\ 10,\ 6,\ 10,\ 17$ \\
  & $12,\ 1,\ 14,\ 16,\ 6,\ 12,\ 16,\ 8,\ 1,\ 10$ & $4,\ 8,\ 13,\ 9,\ 13,\ 5,\ 8,\ 7,\ 2,\ 18$ \\
  & $12,\ 11,\ 2,\ 4,\ 11,\ 11,\ 5,\ 4,\ 5,\ 4$ & $11,\ 12,\ 1,\ 5,\ 3,\ 0,\ 5,\ 15,\ 17,\ 15$ \\
  & $18,\ 4,\ 11,\ 5,\ 0,\ 0,\ 13,\ 11,\ 6,\ 1)$ & $11,\ 5,\ 3,\ 13,\ 12,\ 18,\ 12,\ 10,\ 18,\ 0)$ \\
\midrule
\multirow{2}{*}{\textsc{Prod}, $L=10$}
  & $a = (0,\ 0,\ 2,\ 0,\ 3)$ & \multirow{2}{*}{$Y = (1,\ 1,\ 3,\ 5,\ 3,\ 5,\ 0,\ 0,\ 0,\ 0)$}  \\
  & $b = (0,\ 2,\ 6,\ 3,\ 7)$ &  \\
\bottomrule
\end{tabular}
\label{tab:dataset_samples}
\end{table*}

\subsection{\texorpdfstring{Success rates at $L{=}20$ and $L{=}50$}{Success rates at L=20 and L=50}}
\label{app:success_rate_each_order_full}
\Cref{tab:success_rate_each_order} in the main text reports the success rates at $L{=}20$.
\Cref{tab:success_rate_each_order_full} extends the comparison to $L{=}50$ and shows the same trend across all six tasks: the forward order remains near $100\%$, while the reverse order collapses to single-digit accuracy or below.
The forward-order success rates are slightly higher than at $L{=}20$ in some tasks because longer sequences provide more training tokens per sample under the same dataset size.

\begin{table}[h]
\centering
\small
\setlength{\tabcolsep}{4pt}
\caption{Success rates (\%) of Transformers trained with a fixed target order (forward or reverse), evaluated on validation sets at both $L{=}20$ and $L{=}50$. Across tasks and lengths, the forward order is substantially more learning-friendly. The $L{=}20$ rows are duplicated from \cref{tab:success_rate_each_order} for self-containedness.}
\label{tab:success_rate_each_order_full}
\begin{tabular}{llccc}
\toprule
\multirow{2}{*}{\textbf{Task}} & \multirow{2}{*}{\textbf{Recurrence ($i \ge 2$)}} & \multirow{2}{*}{\textbf{Length}} & \multicolumn{2}{c}{\textbf{Success Rate (\%)}}\\
\cmidrule(lr){4-5}
& & & \textbf{Forward} & \textbf{Reverse}\\
\midrule
\multirow{2}{*}{\textsc{ReLU}}     & \multirow{2}{*}{$y_i = \operatorname{ReLU}(x_i + y_{i-1})$}                                & $L=20$ & \textbf{99.6} & 5.6 \\
                                    &                                                                                            & $L=50$ & \textbf{99.9} & 0.6 \\
\midrule
\multirow{2}{*}{\textsc{Square}}   & \multirow{2}{*}{$y_i = (x_i^2 + y_{i-1}^2)\bmod N$}                                         & $L=20$ & \textbf{100}  & 0.1 \\
                                    &                                                                                            & $L=50$ & \textbf{100}  & 0.0 \\
\midrule
\multirow{2}{*}{\textsc{Triangle}} & \multirow{2}{*}{$y_i = \bigl|((y_{i-1}+x_i)\bmod 2N) - N\bigr|$}                            & $L=20$ & \textbf{100}  & 0.2 \\
                                    &                                                                                            & $L=50$ & \textbf{100}  & 0.0 \\
\midrule
\multirow{2}{*}{\textsc{MLP}}      & \multirow{2}{*}{$y_i = \bigl\lfloor \operatorname{MLP}(x_i, y_{i-1}) \bigr\rfloor \bmod N$} & $L=20$ & \textbf{100}  & 9.4 \\
                                    &                                                                                            & $L=50$ & \textbf{100}  & 0.6 \\
\midrule
\multirow{2}{*}{\textsc{Sine}}     & \multirow{2}{*}{$y_i = \bigl\lfloor A\sin\!\bigl(\tfrac{2\pi}{P}(y_{i-1}+x_i)\bigr) \bigr\rfloor \bmod N$} & $L=20$ & \textbf{100}  & 6.2 \\
                                    &                                                                                            & $L=50$ & \textbf{100}  & 0.3 \\
\midrule
\multirow{2}{*}{\textsc{Cubic}}    & \multirow{2}{*}{$y_i = (y_{i-1}^3 + x_i)\bmod N$}                                          & $L=20$ & \textbf{100}  & 1.6 \\
                                    &                                                                                            & $L=50$ & \textbf{100}  & 0.0 \\
\bottomrule
\end{tabular}
\end{table}

\subsection{Experimental parameters and settings}
\label{app:experimental_settings}

We summarize the specific parameters used for the order-sensitive tasks and the model training in \cref{tab:task_params,tab:training_params}, respectively.

In addition to the parameters listed in the tables, the following common configurations are applied across all experiments.
Optimization is performed using AdamW~\citep{AdamW} with $\beta_1=0.9$ and $\beta_2=0.999$.
We employ a linearly decaying learning rate schedule starting from $5.0 \times 10^{-5}$.
The dropout rate is set to $0.1$, and positional embeddings are randomly initialized and optimized during training.

\begin{table}[h]
\centering
\caption{Parameters for order-sensitive tasks}
\label{tab:task_params}
\begin{tabular}{lccc}
\toprule
\textbf{Task} & \textbf{Modulus ($N$)} & \textbf{Input Range ($x_i$)} & \textbf{Other Constants} \\
\midrule
\textsc{ReLU} & N/A & $\{-9, \dots, 9\}$ & --- \\
\textsc{Square} & $19$ & $\{-9, \dots, 9\}$ & --- \\
\textsc{Cubic} & $19$ & $\{0, \dots, 18\}$ & --- \\
\textsc{Triangle} & $20$ & $\{0, \dots, 19\}$ & Amplitude $N=20$ \\
\textsc{Sine} & $32$ & $\{0, \dots, 31\}$ & $A=100, P=10$ \\
\textsc{MLP} & $32$ & $\{0, \dots, 31\}$ & $d_{\mathrm{hidden}}=64$ \\
\bottomrule
\end{tabular}
\end{table}

\begin{table}[h]
\centering
\small
\setlength{\tabcolsep}{4pt}
\caption{Transformer training configuration.}
\label{tab:training_params}
\begin{tabular}{lccc}
\toprule
\multirow{2}{*}{\textbf{Hyperparameter}} & \multicolumn{3}{c}{\textbf{Phase}} \\
\cmidrule(lr){2-4}
 & \textbf{\shortstack{Exploration\\(Small)}} & \textbf{\shortstack{Final training\\(Large)}} & \textbf{\shortstack{Final training\\(\textsc{Prod})}} \\
\midrule
Layers ($n_{\mathrm{layer}}$) & $1 \sim 4$ & $6$ & $12$ \\
Attention heads ($n_{\mathrm{head}}$) & $1 \sim 4$ & $8$ & $12$\\
Embedding dim ($d_{\mathrm{emb}}$) & $128 \sim 512$ & $512$ & $768$ \\
FFN dim ($d_{\mathrm{ffn}}$) & $512 \sim 2048$ & $2048$ & $3072$\\
Training samples & $10^5$ & $10^5$ & $10^6$\\
Batch size & $32 \sim 128$ & $128$ & $128$\\
Training epochs & $1$ & $10$ & $30$\\
\bottomrule
\end{tabular}
\end{table}

\subsection{Permutation set visualization}\label{app:permutation_vis}
\begin{figure}[h]
  \centering
  \includegraphics[width=1.0\linewidth]{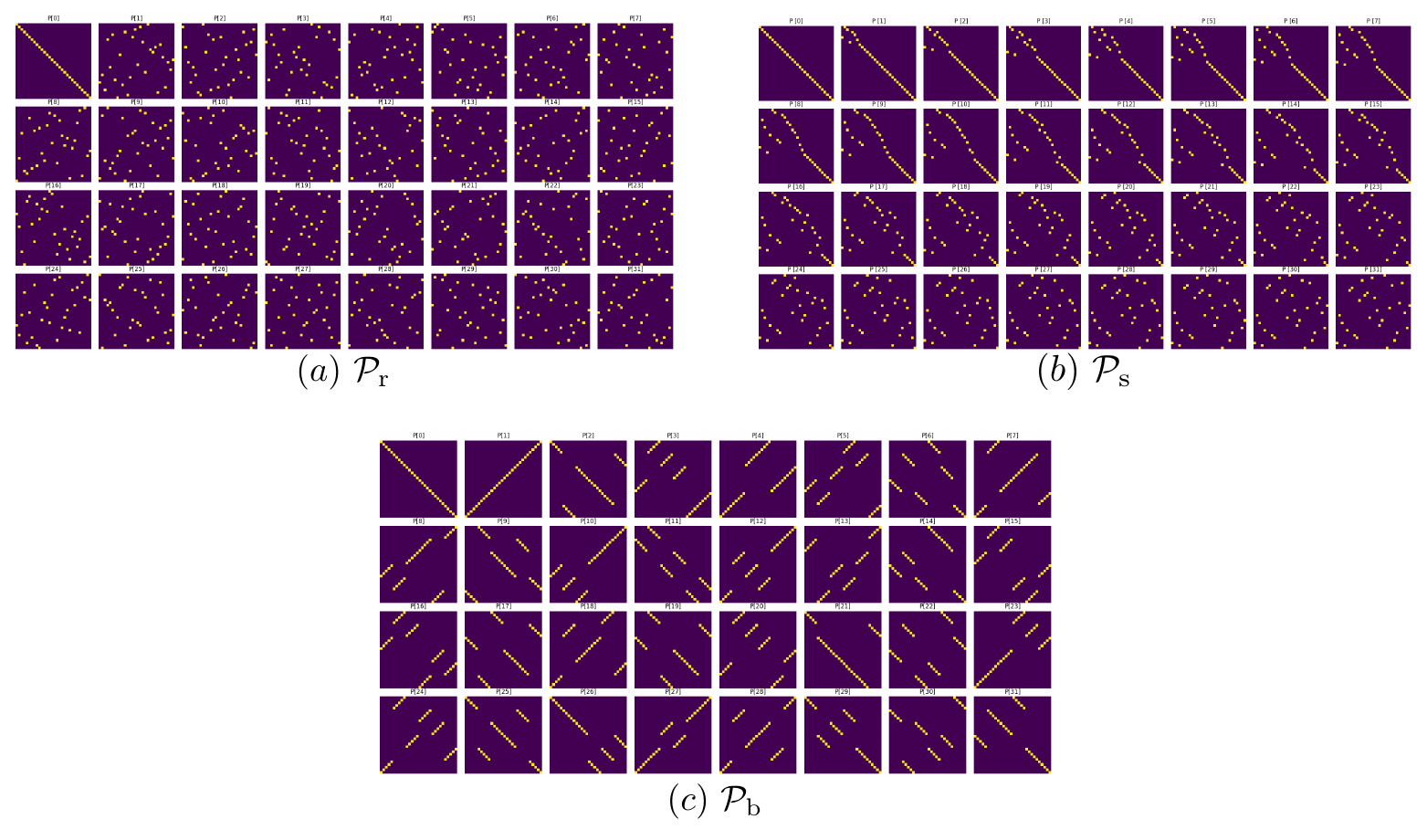}
  \caption{Visualization of the elements in the three permutation sets used in \cref{sec:exp-loss-profiling}. ID=0 corresponds to the forward (learning-friendly) order.}
  \label{fig:permutation_sets_with_forward}
\end{figure}

\begin{figure}[h]
  \centering
  \includegraphics[width=1.0\linewidth]{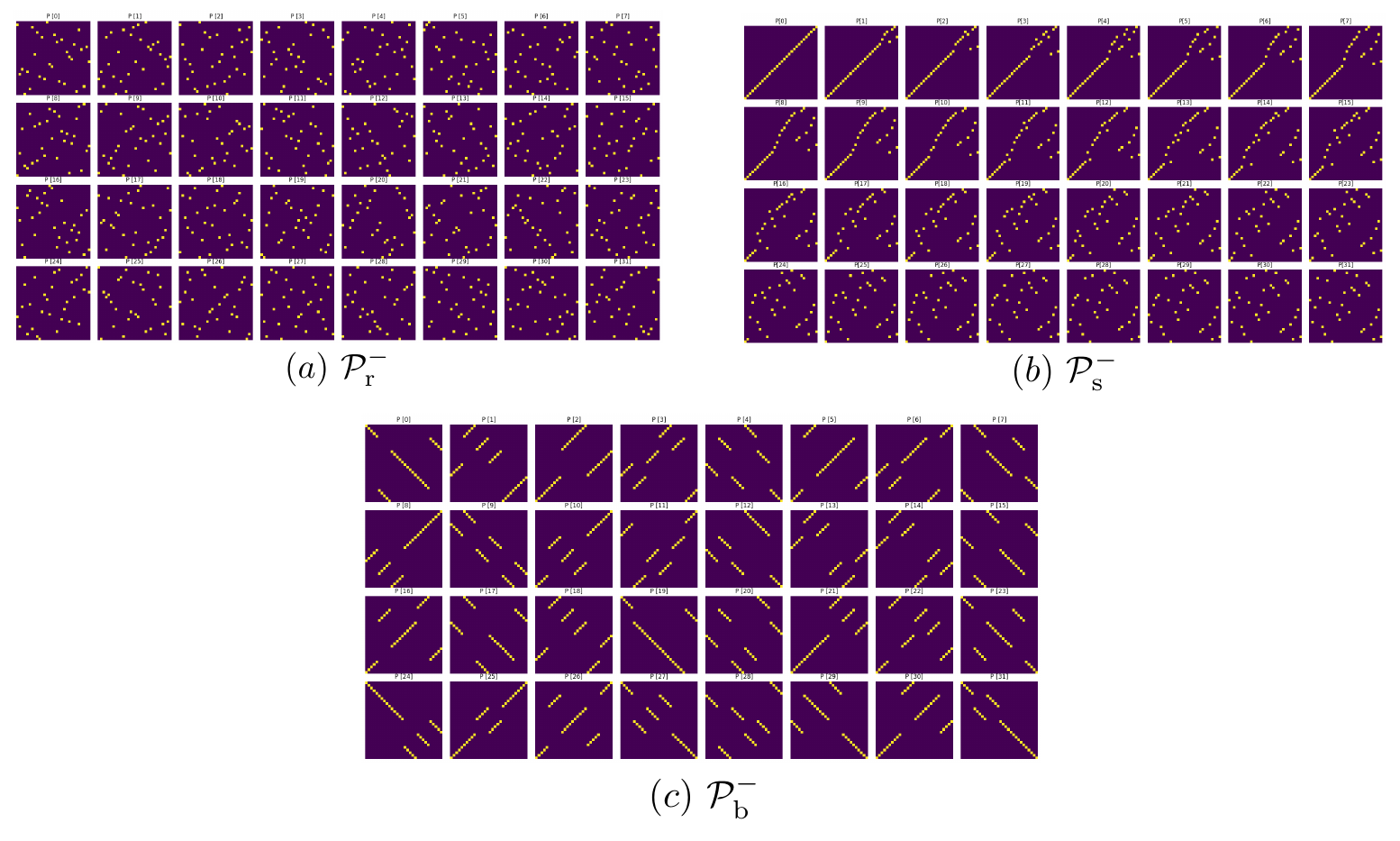}
  \caption{Visualization of the elements in the permutation sets used in \cref{sec:exp-global-local}. Note that the forward order is not included in the initial candidates.}
  \label{fig:permutation_sets_without_forward}
\end{figure}

Next, we visualize the three permutation sets ($\cP_{\mathrm{r}}$, $\cP_{\mathrm{s}}$, $\cP_{\mathrm{b}}$) introduced in \cref{sec:training_setup}.
\Cref{fig:permutation_sets_with_forward} displays the sets used in the loss profiling experiments (\cref{sec:exp-loss-profiling}), which include the forward order (Ground Truth) at ID=0.
\Cref{fig:permutation_sets_without_forward} illustrates the sets used in the global--local search experiments (\cref{sec:exp-global-local}), which do not include the forward order in the initial candidate pool.

\section{Detailed results of loss profiling}
\label{app:all_loss_profiling_results}

In this section, we report the full results of the experiments conducted in \cref{sec:exp-loss-profiling}.
Please refer to \cref{sec:training_setup} and \cref{sec:exp-loss-profiling} for the experimental setup and methodology.
\Cref{tab:profiling_relu_results,tab:profiling_square_results,tab:profiling_sine_results,tab:cubic_results,tab:mlp_results,tab:profiling_triangle_results} present the success rates of loss profiling on the proposed tasks.
In these tables, ``Target Length'' denotes the target sequence length, and ``Permutation'' indicates the specific permutation set used in the experiment.
``Top-$k$ Acc'' indicates the proportion of trials where the optimal forward order is found within the top $k$ ranked candidates.
``Med. Rank'' represents the median rank of the forward order among the 128 candidates based on the validation loss.

\begin{table}[h]
  \centering
  \caption{Performance evaluation on \textsc{ReLU} task.}
  \label{tab:profiling_relu_results}
  \begin{tabular}{lccccc}
    \toprule
    Target Length & Permutation & Top-1 Acc & Top-5 Acc & Top-10 Acc & Med. Rank \\
    \midrule
    \multirow{3}{*}{$L=30$} 
      & $\cP_{\mathrm{r}}$ & 85.19\% & 92.59\% & 96.30\% & 1.00 \\
      & $\cP_{\mathrm{b}}$ & 77.78\% & 96.30\% & 100.00\% & 1.00 \\
      & $\cP_{\mathrm{s}}$ & 96.30\% & 100.00\% & 100.00\% & 1.00 \\
    \cmidrule(lr){1-6}
    \multirow{3}{*}{$L=50$} 
      & $\cP_{\mathrm{r}}$ & 100.00\% & 100.00\% & 100.00\% & 1.00 \\
      & $\cP_{\mathrm{b}}$ & 81.48\% & 100.00\% & 100.00\% & 1.00 \\
      & $\cP_{\mathrm{s}}$ & 62.96\% & 77.78\% & 100.00\% & 1.00 \\
    \cmidrule(lr){1-6}
    \multirow{3}{*}{$L \in \{20, 21\dots, 30\}$} 
      & $\cP_{\mathrm{r}}$ & 62.96\% & 81.48\% & 88.89\% & 1.00 \\
      & $\cP_{\mathrm{b}}$ & 77.78\% & 85.19\% & 88.89\% & 1.00 \\
      & $\cP_{\mathrm{s}}$ & 70.37\% & 96.30\% & 100.00\% & 1.00 \\
    \cmidrule(lr){1-6}
    \multirow{3}{*}{$L \in \{40, 41\dots, 50\}$} 
      & $\cP_{\mathrm{r}}$ & 96.30\% & 100.00\% & 100.00\% & 1.00 \\
      & $\cP_{\mathrm{b}}$ & 66.67\% & 100.00\% & 100.00\% & 1.00 \\
      & $\cP_{\mathrm{s}}$ & 66.67\% & 85.19\% & 96.30\% & 1.00 \\
    \bottomrule
  \end{tabular}
\end{table}

\begin{table}[h]
  \centering
  \caption{Performance evaluation on \textsc{Square} task.}
  \label{tab:profiling_square_results}
  \begin{tabular}{lccccc}
    \toprule
    Target Length & Permutation & Top-1 Acc & Top-5 Acc & Top-10 Acc & Med. Rank \\
    \midrule
    \multirow{3}{*}{$L=30$} 
      & $\cP_{\mathrm{r}}$ & 77.78\% & 85.19\% & 85.19\% & 4.00 \\
      & $\cP_{\mathrm{b}}$ & 92.31\% & 100.00\% & 100.00\% & 1.00 \\
      & $\cP_{\mathrm{s}}$ & 55.56\% & 66.67\% & 100.00\% & 1.00 \\
    \cmidrule(lr){1-6}
    \multirow{3}{*}{$L=50$} 
      & $\cP_{\mathrm{r}}$ & 96.30\% & 100.00\% & 100.00\% & 1.00 \\
      & $\cP_{\mathrm{b}}$ & 62.96\% & 66.67\% & 66.67\% & 1.00 \\
      & $\cP_{\mathrm{s}}$ & 33.33\% & 40.74\% & 96.30\% & 6.00 \\
    \cmidrule(lr){1-6}
    \multirow{3}{*}{$L \in \{20, 21\dots, 30\}$} 
      & $\cP_{\mathrm{r}}$ & 70.37\% & 92.59\% & 96.30\% & 1.00 \\
      & $\cP_{\mathrm{b}}$ & 96.30\% & 100.00\% & 100.00\% & 1.00 \\
      & $\cP_{\mathrm{s}}$ & 11.11\% & 11.11\% & 81.48\% & 9.00 \\
    \cmidrule(lr){1-6}
    \multirow{3}{*}{$L \in \{40, 41\dots, 50\}$} 
      & $\cP_{\mathrm{r}}$ & 100.00\% & 100.00\% & 100.00\% & 1.00 \\
      & $\cP_{\mathrm{b}}$ & 66.67\% & 66.67\% & 66.67\% & 1.00 \\
      & $\cP_{\mathrm{s}}$ & 33.33\% & 44.44\% & 92.59\% & 6.00 \\
    \bottomrule
  \end{tabular}
\end{table}

\begin{table}[h]
  \centering
  \caption{Performance evaluation on \textsc{Sine} task.}
  \label{tab:profiling_sine_results}
  \begin{tabular}{lccccc}
    \toprule
    Target Length & Permutation & Top-1 Acc & Top-5 Acc & Top-10 Acc & Med. Rank \\
    \midrule
    \multirow{3}{*}{$L=30$} 
      & $\cP_{\mathrm{r}}$ & 37.04\% & 44.44\% & 48.15\% & 14.00 \\
      & $\cP_{\mathrm{b}}$ & 88.89\% & 92.59\% & 96.30\% & 1.00 \\
      & $\cP_{\mathrm{s}}$ & 11.11\% & 66.67\% & 70.37\% & 3.00 \\
    \cmidrule(lr){1-6}
    \multirow{3}{*}{$L=50$} 
      & $\cP_{\mathrm{r}}$ & 55.56\% & 77.78\% & 77.78\% & 1.00 \\
      & $\cP_{\mathrm{b}}$ & 33.33\% & 51.85\% & 55.56\% & 2.00 \\
      & $\cP_{\mathrm{s}}$ & 59.26\% & 59.26\% & 59.26\% & 1.00 \\
    \cmidrule(lr){1-6}
    \multirow{3}{*}{$L \in \{20, 21\dots, 30\}$} 
      & $\cP_{\mathrm{r}}$ & 18.52\% & 25.93\% & 44.44\% & 11.00 \\
      & $\cP_{\mathrm{b}}$ & 96.30\% & 100.00\% & 100.00\% & 1.00 \\
      & $\cP_{\mathrm{s}}$ & 0.00\% & 18.52\% & 66.67\% & 7.00 \\
    \cmidrule(lr){1-6}
    \multirow{3}{*}{$L \in \{40, 41\dots, 50\}$} 
      & $\cP_{\mathrm{r}}$ & 62.96\% & 66.67\% & 77.78\% & 1.00 \\
      & $\cP_{\mathrm{b}}$ & 48.15\% & 66.67\% & 70.37\% & 2.00 \\
      & $\cP_{\mathrm{s}}$ & 48.15\% & 55.56\% & 55.56\% & 2.00 \\
    \bottomrule
  \end{tabular}
\end{table}

\begin{table}[h]
  \centering
  \caption{Performance evaluation on \textsc{Cubic} task.}
  \label{tab:cubic_results}
  \begin{tabular}{lccccc}
    \toprule
    Target Length & Permutation & Top-1 Acc & Top-5 Acc & Top-10 Acc & Med. Rank \\
    \midrule
    \multirow{3}{*}{$L=30$} 
      & $\cP_{\mathrm{r}}$ & 18.52\% & 22.22\% & 22.22\% & 83.00 \\
      & $\cP_{\mathrm{b}}$ & 55.56\% & 85.19\% & 85.19\% & 1.00 \\
      & $\cP_{\mathrm{s}}$ & 0.00\% & 0.00\% & 59.26\% & 8.00 \\
    \cmidrule(lr){1-6}
    \multirow{3}{*}{$L=50$} 
      & $\cP_{\mathrm{r}}$ & 22.22\% & 44.44\% & 59.26\% & 8.00 \\
      & $\cP_{\mathrm{b}}$ & 55.56\% & 66.67\% & 66.67\% & 1.00 \\
      & $\cP_{\mathrm{s}}$ & 0.00\% & 0.00\% & 11.11\% & 25.00 \\
    \cmidrule(lr){1-6}
    \multirow{3}{*}{$L \in \{20, 21\dots, 30\}$} 
      & $\cP_{\mathrm{r}}$ & 3.70\% & 7.41\% & 11.11\% & 55.00 \\
      & $\cP_{\mathrm{b}}$ & 74.07\% & 100.00\% & 100.00\% & 1.00 \\
      & $\cP_{\mathrm{s}}$ & 0.00\% & 0.00\% & 48.15\% & 15.00 \\
    \cmidrule(lr){1-6}
    \multirow{3}{*}{$L \in \{40, 41\dots, 50\}$} 
      & $\cP_{\mathrm{r}}$ & 0.00\% & 22.22\% & 25.93\% & 39.00 \\
      & $\cP_{\mathrm{b}}$ & 74.07\% & 77.78\% & 77.78\% & 1.00 \\
      & $\cP_{\mathrm{s}}$ & 0.00\% & 0.00\% & 0.00\% & 31.00 \\
    \bottomrule
  \end{tabular}
\end{table}

\begin{table}[h]
  \centering
  \caption{Performance evaluation on \textsc{MLP} task.}
  \label{tab:mlp_results}
  \begin{tabular}{lccccc}
    \toprule
    Target Length & Permutation & Top-1 Acc & Top-5 Acc & Top-10 Acc & Med. Rank \\
    \midrule
    \multirow{3}{*}{$L=30$} 
      & $\cP_{\mathrm{r}}$ & 22.22\% & 25.93\% & 37.04\% & 26.00 \\
      & $\cP_{\mathrm{b}}$ & 100.00\% & 100.00\% & 100.00\% & 1.00 \\
      & $\cP_{\mathrm{s}}$ & 0.00\% & 18.52\% & 85.19\% & 7.00 \\
    \cmidrule(lr){1-6}
    \multirow{3}{*}{$L=50$} 
      & $\cP_{\mathrm{r}}$ & 85.19\% & 96.30\% & 96.30\% & 1.00 \\
      & $\cP_{\mathrm{b}}$ & 88.89\% & 100.00\% & 100.00\% & 1.00 \\
      & $\cP_{\mathrm{s}}$ & 0.00\% & 0.00\% & 0.00\% & 25.00 \\
    \cmidrule(lr){1-6}
    \multirow{3}{*}{$L \in \{20, 21\dots, 30\}$} 
      & $\cP_{\mathrm{r}}$ & 18.52\% & 25.93\% & 40.74\% & 13.00 \\
      & $\cP_{\mathrm{b}}$ & 100.00\% & 100.00\% & 100.00\% & 1.00 \\
      & $\cP_{\mathrm{s}}$ & 25.93\% & 96.30\% & 100.00\% & 2.00 \\
    \cmidrule(lr){1-6}
    \multirow{3}{*}{$L \in \{40, 41\dots, 50\}$} 
      & $\cP_{\mathrm{r}}$ & 18.52\% & 29.63\% & 51.85\% & 10.00 \\
      & $\cP_{\mathrm{b}}$ & 66.67\% & 85.19\% & 88.89\% & 1.00 \\
      & $\cP_{\mathrm{s}}$ & 37.04\% & 37.04\% & 37.04\% & 15.00 \\
    \bottomrule
  \end{tabular}
\end{table}

\begin{table}[h]
  \centering
  \caption{Performance evaluation on \textsc{Triangle} task.}
  \label{tab:profiling_triangle_results}
  \begin{tabular}{lccccc}
    \toprule
    Target Length & Permutation & Top-1 Acc & Top-5 Acc & Top-10 Acc & Med. Rank \\
    \midrule
    \multirow{3}{*}{$L=30$} 
      & $\cP_{\mathrm{r}}$ & 48.15\% & 51.85\% & 51.85\% & 2.00 \\
      & $\cP_{\mathrm{b}}$ & 96.30\% & 100.00\% & 100.00\% & 1.00 \\
      & $\cP_{\mathrm{s}}$ & 59.26\% & 85.19\% & 88.89\% & 1.00 \\
    \cmidrule(lr){1-6}
    \multirow{3}{*}{$L=50$} 
      & $\cP_{\mathrm{r}}$ & 96.30\% & 100.00\% & 100.00\% & 1.00 \\
      & $\cP_{\mathrm{b}}$ & 66.67\% & 66.67\% & 66.67\% & 1.00 \\
      & $\cP_{\mathrm{s}}$ & 37.04\% & 37.04\% & 37.04\% & 12.00 \\
    \cmidrule(lr){1-6}
    \multirow{3}{*}{$L \in \{20, 21\dots, 30\}$} 
      & $\cP_{\mathrm{r}}$ & 48.15\% & 62.96\% & 85.19\% & 2.00 \\
      & $\cP_{\mathrm{b}}$ & 100.00\% & 100.00\% & 100.00\% & 1.00 \\
      & $\cP_{\mathrm{s}}$ & 7.41\% & 59.26\% & 85.19\% & 4.00 \\
    \cmidrule(lr){1-6}
    \multirow{3}{*}{$L \in \{40, 41, \dots, 50\}$} 
      & $\cP_{\mathrm{r}}$ & 70.37\% & 85.19\% & 92.59\% & 1.00 \\
      & $\cP_{\mathrm{b}}$ & 74.07\% & 81.48\% & 88.89\% & 1.00 \\
      & $\cP_{\mathrm{s}}$ & 33.33\% & 33.33\% & 37.04\% & 13.00 \\
    \bottomrule
  \end{tabular}
\end{table}

\FloatBarrier
\section{Correlation between loss-profiling ranks and retraining accuracy}
\label{app:profiling_retrain_correlation}

\subsection{Identification accuracy across model dimensions}
\label{app:loss_profiling_dim_ablation}
\begin{figure}[h]
    \centering
    \includegraphics[width=1.0\linewidth]{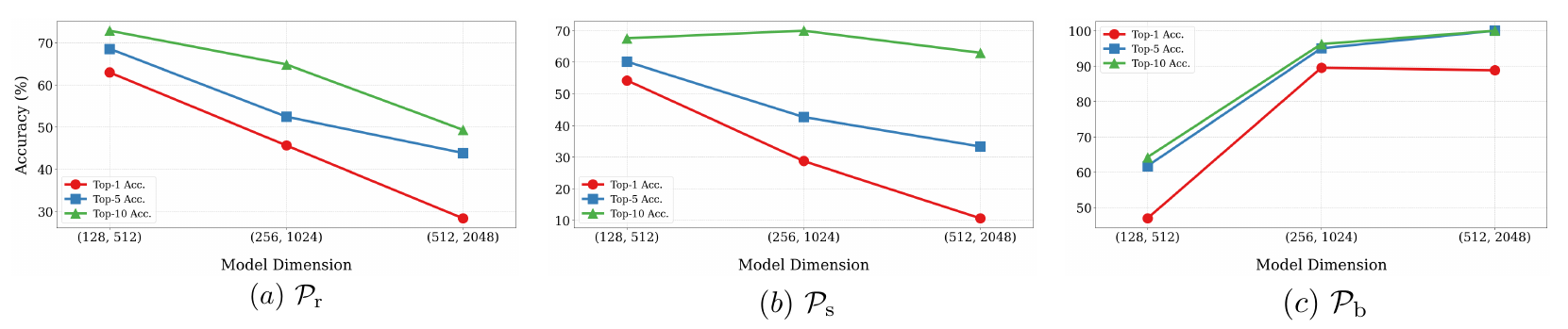}
    \caption{Identification accuracy of the forward order across various model dimensions, defined by pairs of $(d_{\mathrm{emb}}, d_{\mathrm{ffn}})$. Subplots (a), (b), and (c) show the results for the initialization sets $\cP_{\mathrm{r}}$, $\cP_{\mathrm{s}}$, and $\cP_{\mathrm{b}}$, respectively.}
    \label{fig:loss-profiling-acc-by-dim}
\end{figure}

\Cref{fig:loss-profiling-acc-by-dim} presents the impact of model dimensions $(d_{\mathrm{emb}}, d_{\mathrm{ffn}})$ on the top-1 identification accuracy of loss profiling, complementing the brief summary in \cref{sec:exp-loss-profiling}.
For $\cP_{\mathrm{r}}$ and $\cP_{\mathrm{s}}$, smaller model dimensions yield higher identification accuracy, suggesting that models with limited expressivity are more sensitive to the difficulty of the order and adapt more easily to learning-friendly sequences.
Conversely, for $\cP_{\mathrm{b}}$, higher model dimensions correlate with better identification accuracy: when candidates are locally similar (as in block permutations), a larger capacity is required to distinguish global structural differences.

\subsection{Correlation analysis}
\label{app:profiling_retrain_correlation_main}

The loss-profiling results in \cref{app:all_loss_profiling_results} evaluate whether the known forward order is ranked near the top.
Here, we additionally ask whether the \emph{entire} profiling ranking is predictive of the accuracy obtained after retraining with each candidate order fixed.
For each task and target length, we fix the retraining accuracies from a reference run and pair them with the profiling loss of the same permutation index.
We analyze \textsc{Cubic}, \textsc{MLP}, and \textsc{ReLU} at length $L=30$.
The profiling side contains the same 27 GPT-2 configurations used in \cref{sec:exp-loss-profiling}: $n_{\mathrm{layer}} \in \{1,2,4\}$, $n_{\mathrm{head}} \in \{1,2,4\}$, and $(d_{\mathrm{emb}}, d_{\mathrm{ffn}}) \in \{(128,512),(256,1024),(512,2048)\}$.

We report four metrics.
Spearman correlation is computed between the profiling rank and the retraining-accuracy rank, so larger values indicate stronger agreement.
Pearson correlation is computed between the raw profiling loss and the raw test accuracy; a useful profiling signal should produce a negative value because lower profiling loss should correspond to higher accuracy.
Top-5 overlap is the fraction of candidates shared by the profiling top-5 and the retraining-accuracy top-5.
Mean absolute rank gap is the average absolute difference between the two ranks, where smaller values are better.

\begin{table}[h]
  \centering
  \small
  \setlength{\tabcolsep}{5pt}
  \caption{Correlation between loss-profiling rankings and retraining accuracy rankings at $L=30$. ``All'' aggregates the three tasks and all 27 profiling configurations per task.}
  \label{tab:profiling_retrain_correlation_task}
  \begin{tabular}{lrrrrr}
    \toprule
    \textbf{Task} & \textbf{Runs} & \textbf{Spearman} & \textbf{Pearson} & \textbf{Top-5 overlap} & \textbf{Rank gap} \\
    \midrule
    All & 81 & 0.6739 & -0.5692 & 0.5704 & 17.8974 \\
    \textsc{ReLU} & 27 & 0.9241 & -0.8090 & 0.6889 & 10.0877 \\
    \textsc{MLP} & 27 & 0.8977 & -0.8058 & 0.5852 & 12.4363 \\
    \textsc{Cubic} & 27 & 0.1999 & -0.0928 & 0.4370 & 31.1681 \\
    \bottomrule
  \end{tabular}
\end{table}

\begin{table}[h]
  \centering
  \small
  \setlength{\tabcolsep}{6pt}
  \caption{Correlation metrics grouped by profiling-model embedding dimension. Each row aggregates all three tasks and the corresponding profiling configurations.}
  \label{tab:profiling_retrain_correlation_dim}
  \begin{tabular}{lrrrrr}
    \toprule
    \textbf{Dim.} & \textbf{Runs} & \textbf{Spearman} & \textbf{Pearson} & \textbf{Top-5 overlap} & \textbf{Rank gap} \\
    \midrule
    $D=128$ & 27 & 0.6094 & -0.5281 & 0.2815 & 18.8854 \\
    $D=256$ & 27 & 0.7060 & -0.5852 & 0.7111 & 17.3938 \\
    $D=512$ & 27 & 0.7064 & -0.5944 & 0.7185 & 17.4129 \\
    \bottomrule
  \end{tabular}
\end{table}

\Cref{tab:profiling_retrain_correlation_task} shows that the profiling ranking strongly agrees with the retraining-accuracy ranking on \textsc{ReLU} and \textsc{MLP}.
\textsc{Cubic} is the main exception: the correlation is weak, consistent with the difficulty of obtaining high retraining accuracy on this task even under favorable orders.
Thus, loss profiling is best interpreted as a screening signal for candidate orders rather than a perfect predictor of final accuracy.
\Cref{tab:profiling_retrain_correlation_dim} further shows that $D=256$ and $D=512$ yield substantially higher top-5 overlap than $D=128$, suggesting that moderately larger profiling models can be useful when the goal is to select a small pool of promising orders for retraining.

\FloatBarrier
\section{Detailed results of global--local pipeline}
\label{app:global_local_details}

In this section, we present the detailed experimental results from the global--local pipeline described in \cref{sec:exp-global-local}.
\Cref{tab:global_local_ablation_acc,tab:global_local_ablation_orders_l10,tab:global_local_structured_orders_l20,tab:global_local_structured_orders_l40} report an ablation comparing the global stage alone with the full global--local pipeline.
\Cref{tab:reverse-baseline-extended} reports the success rates of final training when the target sequence is fixed to the reverse order, providing a baseline at every length used in the discovery experiments.
\Cref{tab:global_local_discovered_orders} details the actual orders discovered by the pipeline for each task and target length.
``Depth'' indicates the hierarchy level $K$ reached in the global stage.
The orders are listed relative to the forward sequence (i.e., indices starting from 0 represent the forward order).
Orders that successfully recovered the forward sequence are highlighted in bold.
We also provide the runtime analysis of the exploration and final training phases in \cref{tab:runtime_analysis_appendix}.
The table reports the breakdown of exploration time into global and local stages, and the percentage of time spent on model training during exploration.
For non-\textsc{Prod} order-sensitive tasks, runtime is almost identical across tasks because the training setup (e.g., training steps and input length) is shared; therefore, we report representative timings measured on \textsc{ReLU}.
For \textsc{Prod}, exploration remains shorter than a single full training run across the settings in \cref{tab:runtime_analysis_appendix}. For the other order-sensitive tasks, exploration is not always shorter than a single run, but it still pays off because exhaustive order ranking would require many full retraining runs, whereas loss profiling screens candidates in a single exploration pass.

\subsection{Ablation of the local stage}
\label{app:global_local_ablation}

To quantify the contribution of the local stage, we compare the order obtained after the global stage alone with the order obtained after running the local stage as well.
For each discovered order, we retrain the model with that order fixed and report the resulting sequence-level exact-match accuracy.
\Cref{tab:global_local_ablation_acc} shows that the local stage can substantially improve performance, especially on \textsc{ReLU} and on the $L=10$ \textsc{Square} and \textsc{Triangle} settings.
The effect is task-dependent: on \textsc{Cubic}, and on \textsc{Square} at $L=13$, the local stage does not improve the final accuracy, and the final accuracy remains low.
This suggests that local refinement is useful when the global stage has found a promising coarse structure, but it does not by itself solve settings where the overall search remains difficult.

\begin{table}[h]
\centering
\small
\setlength{\tabcolsep}{4pt}
\caption{Ablation of the local stage across all tasks with available $L=10$ and $L=13$ results. Accuracy is measured after retraining with the order discovered by either the global stage alone or the full global--local pipeline. The change columns report Global+Local minus Global-only in percentage points.}
\label{tab:global_local_ablation_acc}
\begin{tabular}{lrrrrrr}
\toprule
\multirow{2}{*}{\textbf{Task}} &
\multicolumn{3}{c}{$L=10$} &
\multicolumn{3}{c}{$L=13$} \\
\cmidrule(lr){2-4}\cmidrule(lr){5-7}
& \textbf{Global} & \textbf{Global+Local} & \textbf{Change} &
\textbf{Global} & \textbf{Global+Local} & \textbf{Change} \\
\midrule
\textsc{ReLU}     & 30.9 & 99.6  & +68.7 & 6.6  & 99.2 & +92.6 \\
\textsc{Square}   & 49.7 & 100.0 & +50.3 & 8.1  & 5.6  & -2.5 \\
\textsc{Triangle} & 11.4 & 100.0 & +88.6 & 2.9  & 7.4  & +4.5 \\
\textsc{MLP}      & 22.1 & 38.2  & +16.1 & 3.1  & 10.8 & +7.7 \\
\textsc{Sine}     & 37.9 & 54.2  & +16.3 & 23.2 & 26.3 & +3.1 \\
\textsc{Cubic}    & 6.4  & 0.3   & -6.1  & 0.7  & 0.2  & -0.5 \\
\textsc{Prod}     & 100.0 & 100.0 & +0.0  & ---  & ---  & --- \\
\bottomrule
\end{tabular}
\end{table}

\Cref{tab:global_local_ablation_orders_l10} gives the $L=10$ cases where both the global-stage order and the final local-stage order are available.
On \textsc{ReLU}, the local stage turns a coarse, mostly reversed block structure into the forward order.
On \textsc{Square} and \textsc{Prod}, the local stage either recovers or preserves the forward order.
On \textsc{MLP}, it produces a different partially structured order and yields a smaller accuracy gain.
On \textsc{Cubic}, the local stage moves to the reverse order, matching the accuracy degradation in \cref{tab:global_local_ablation_acc}.
\Cref{tab:global_local_structured_orders_l20,tab:global_local_structured_orders_l40} further show the longer structured-initialization cases.
These longer runs do not include a local stage, but the global stage often recovers a coarse order close to the forward sequence under $\cP_{\mathrm{b}}^-$ or $\cP_{\mathrm{s}}^-$.

\begin{table}[h]
\centering
\footnotesize
\setlength{\tabcolsep}{3pt}
\caption{Discovered orders before and after local refinement at $L=10$ with $\cP_{\mathrm{r}}^-$ initialization. Orders are shown as indices relative to the forward sequence.}
\label{tab:global_local_ablation_orders_l10}
\begin{tabular}{lcc>{\raggedright\arraybackslash}p{0.20\textwidth}>{\raggedright\arraybackslash}p{0.20\textwidth}r}
\toprule
\textbf{Task} & \textbf{Initialization} & \textbf{Target Length} & \textbf{Global-stage order} & \textbf{After local stage} & \textbf{Accuracy change} \\
\midrule
\textsc{ReLU}   & $\cP_{\mathrm{r}}^-$ & $L=10$ & [6, 7, 8, 9, 5, 4, 2, 3, 1, 0] & [0, 1, 2, 3, 4, 5, 6, 7, 8, 9] & $30.9 \to 99.6$ \\
\textsc{Square} & $\cP_{\mathrm{r}}^-$ & $L=10$ & [9, 8, 7, 6, 5, 4, 3, 2, 1, 0] & [0, 1, 2, 3, 4, 5, 6, 7, 8, 9] & $49.7 \to 100.0$ \\
\textsc{MLP}    & $\cP_{\mathrm{r}}^-$ & $L=10$ & [9, 8, 5, 3, 2, 1, 0, 4, 6, 7] & [0, 1, 3, 2, 4, 9, 8, 7, 6, 5] & $22.1 \to 38.2$ \\
\textsc{Cubic}  & $\cP_{\mathrm{r}}^-$ & $L=10$ & [0, 1, 5, 6, 7, 8, 9, 2, 3, 4] & [9, 8, 7, 6, 5, 4, 3, 2, 1, 0] & $6.4 \to 0.3$ \\
\textsc{Prod}   & $\cP_{\mathrm{r}}^-$ & $L=10$ & [0, 1, 2, 3, 4, 5, 6, 7, 8, 9] & [0, 1, 2, 3, 4, 5, 6, 7, 8, 9] & $100.0 \to 100.0$ \\
\bottomrule
\end{tabular}
\end{table}

\begin{table}[h]
\centering
\footnotesize
\setlength{\tabcolsep}{2pt}
\caption{Additional global-stage orders at $L=20$ from structured initializations $\cP_{\mathrm{s}}^-$ and $\cP_{\mathrm{b}}^-$. The local stage was not run in these longer settings.}
\label{tab:global_local_structured_orders_l20}
\begin{tabular}{llc>{\raggedright\arraybackslash}p{0.60\textwidth}}
\toprule
\textbf{Task} & \textbf{Initialization} & \textbf{Target Length} & \textbf{Global-stage order} \\
\midrule
\textsc{ReLU}  & $\cP_{\mathrm{s}}^-$ & $L=20$ & [0, 1, 2, 3, 4, 5, 6, 7, 8, 9, 10, 11, 12, 13, 14, 15, 16, 18, 19] \\
\textsc{ReLU}  & $\cP_{\mathrm{b}}^-$ & $L=20$ & [0, 1, 2, 3, 4, 5, 6, 7, 8, 9, 10, 11, 12, 13, 14, 15, 16, 18, 19] \\
\textsc{MLP}   & $\cP_{\mathrm{s}}^-$ & $L=20$ & [0, 1, 2, 3, 4, 5, 6, 7, 8, 9, 10, 11, 12, 13, 14, 19, 18, 17, 16, 15] \\
\textsc{MLP}   & $\cP_{\mathrm{b}}^-$ & $L=20$ & [0, 1, 2, 3, 4, 5, 6, 8, 9, 10, 11, 12, 13, 14, 15, 16, 18, 19] \\
\textsc{Cubic} & $\cP_{\mathrm{s}}^-$ & $L=20$ & [0, 1, 2, 3, 4, 5, 6, 8, 9, 10, 11, 12, 13, 14, 15, 16, 17, 18, 19] \\
\textsc{Cubic} & $\cP_{\mathrm{b}}^-$ & $L=20$ & [19, 18, 17, 16, 15, 14, 13, 12, 11, 10, 9, 8, 7, 6, 5, 4, 3, 2, 1, 0] \\
\bottomrule
\end{tabular}
\end{table}

\begin{table}[h]
\centering
\footnotesize
\setlength{\tabcolsep}{2pt}
\caption{Additional global-stage orders at $L=40$ from structured initialization $\cP_{\mathrm{b}}^-$. The local stage was not run in these longer settings.}
\label{tab:global_local_structured_orders_l40}
\begin{tabular}{llc>{\raggedright\arraybackslash}p{0.59\textwidth}}
\toprule
\textbf{Task} & \textbf{Initialization} & \textbf{Target Length} & \textbf{Global-stage order} \\
\midrule
\textsc{ReLU} & $\cP_{\mathrm{b}}^-$ & $L=40$ & [0, 1, 2, 3, 4, 5, 6, 7, 8, 9, 10, 11, 12, 13, 14, 15, 16, 18, 19, 20, 21, 22, 23, 24, 25, 26, 28, 29, 30, 31, 32, 33, 34, 35, 36, 38, 39] \\
\textsc{Square} & $\cP_{\mathrm{b}}^-$ & $L=40$ & [39, 38, 37, 36, 35, 34, 33, 32, 31, 30, 19, 18, 17, 16, 15, 14, 13, 12, 11, 10, 29, 28, 27, 26, 25, 24, 23, 22, 21, 20, 9, 8, 7, 6, 5, 4, 3, 2, 1, 0] \\
\textsc{Triangle} & $\cP_{\mathrm{b}}^-$ & $L=40$ & [0, 1, 2, 3, 4, 5, 6, 7, 8, 9, 10, 11, 12, 13, 14, 15, 16, 17, 18, 19, 39, 38, 37, 36, 35, 34, 33, 32, 31, 30, 29, 28, 27, 26, 25, 24, 23, 22, 21, 20] \\
\textsc{MLP} & $\cP_{\mathrm{b}}^-$ & $L=40$ & [0, 1, 2, 3, 4, 5, 6, 7, 8, 9, 10, 11, 12, 13, 14, 15, 16, 17, 18, 19, 20, 21, 22, 23, 24, 25, 26, 27, 28, 29, 30, 31, 32, 33, 34, 35, 36, 37, 38, 39] \\
\textsc{Sine} & $\cP_{\mathrm{b}}^-$ & $L=40$ & [39, 38, 37, 36, 35, 34, 33, 32, 31, 30, 19, 18, 17, 16, 15, 14, 13, 12, 11, 10, 29, 28, 27, 26, 25, 24, 23, 22, 21, 20, 9, 8, 7, 6, 5, 4, 3, 2, 1, 0] \\
\textsc{Cubic} & $\cP_{\mathrm{b}}^-$ & $L=40$ & [0, 1, 2, 3, 4, 5, 6, 7, 8, 9, 10, 11, 12, 13, 14, 15, 16, 17, 18, 19, 39, 38, 37, 36, 35, 34, 33, 32, 31, 30, 29, 28, 27, 26, 25, 24, 23, 22, 21, 20] \\
\bottomrule
\end{tabular}
\end{table}

\FloatBarrier

\begin{table}[h]
\centering
\small
\setlength{\tabcolsep}{6pt}
\caption{Success rates of final training when the target sequence is fixed to the reverse order, evaluated at the same lengths as the discovery experiments in \cref{tab:global-local-main-results}. The reverse order yields near-zero accuracy on all tasks, confirming that the discovered orders provide the substantial gains reported in \cref{sec:exp-global-local}.}
\label{tab:reverse-baseline-extended}
\begin{tabular}{lcccccc}
\toprule
\textbf{Target Length} & \textsc{ReLU} & \textsc{Square} & \textsc{Triangle} & \textsc{MLP} & \textsc{Sine} & \textsc{Cubic} \\
\midrule
$L=10$ & 2.7 & 6.5 & 3.5 & 10.4 & 21.8 & 0.6 \\
$L=13$ & 1.9 & 3.2 & 3.7 & 8.2  & 15.2 & 1.4 \\
$L=20$ & 5.6 & 0.1 & 0.2 & 9.4  & 6.2  & 1.6 \\
$L=30$ & 1.1 & 0.1 & 0.2 & 3.8  & 1.1  & 0.0 \\
$L=40$ & 0.9 & 0.0 & 0.0 & 0.8  & 0.7  & 0.0 \\
\bottomrule
\end{tabular}
\end{table}

\FloatBarrier
\begin{table*}[h]
\centering
\caption{The orders discovered by the proposed method in its global and local stages.
Depth denotes the hierarchy level $K$ reached in the global stage.
Each order is listed relative to the forward sequence;
when the list starts at 0, the forward order has been recovered.
Forward orders identified at a given stage are highlighted in bold.}
\label{tab:global_local_discovered_orders}
\resizebox{\textwidth}{!}{%
\begin{tabular}{c l l l l}
\toprule
\textbf{Task} & \textbf{Target Length} & \textbf{Depth} & \textbf{Order after global stage} & \textbf{Discovered final order} \\
\midrule

\multirow{7}{*}{\textsc{ReLU}}
& $L=7$  & $K=4$ & $[6, 0, 5, 2, 3, 4, 1]$ & $[2, 3, 4, 5, 0, 6, 1]$ \\
& $L=8$  & $K=4$ & $[0, 2, 1, 3, 4, 5, 6, 7]$ & $[\mathbf{0, 1, 2, 3, 4, 5, 6, 7}]$ \\
& $L=9$  & $K=5$ & $[\mathbf{0, 1, 2, 3, 4, 5, 6, 7, 8}]$ & $[\mathbf{0, 1, 2, 3, 4, 5, 6, 7, 8}]$ \\
& $L=10$ & $K=6$ & $[6, 7, 8, 9, 5, 4, 2, 3, 1, 0]$ & $[4, 5, 6, 7, 8, 9, 0, 1, 2, 3]$ \\
& $L=11$ & $K=6$ & $[8, 9, 10, 7, 6, 5, 4, 3, 2, 1, 0]$ & $\mathbf{[0, 1, 2, 3, 4, 5, 6, 7, 8, 9, 10]}$ \\
& $L=12$ & $K=6$ & $[6, 7, 8, 9, 10, 11, 5, 4, 2, 3, 1, 0]$ & $[1, 2, 3, 4, 0, 5, 6, 7, 8, 9, 10, 11]$ \\
& $L=13$ & $K=6$ & $[11, 12, 10, 9, 8, 7, 6, 5, 4, 2, 3, 1, 0]$ & $\mathbf{[0, 1, 2, 3, 4, 5, 6, 7, 8, 9, 10, 11, 12]}$ \\
\midrule

\multirow{7}{*}{\textsc{Square}}
& $L=7$  & $K=4$ & $\mathbf{[0, 1, 2, 3, 4, 5, 6]}$ & $\mathbf{[0, 1, 2, 3, 4, 5, 6]}$ \\
& $L=8$  & $K=4$ & $[1, 2, 4, 5, 0, 6, 7, 3]$ & $[1, 2, 4, 5, 0, 6, 7, 3]$ \\
& $L=9$  & $K=5$ & $\mathbf{[0, 1, 2, 3, 4, 5, 6, 7, 8}]$ & $\mathbf{[0, 1, 2, 3, 4, 5, 6, 7, 8}]$ \\
& $L=10$ & $K=6$ & $[9, 8, 7, 6, 5, 4, 3, 2, 1, 0]$ & $\mathbf{[0, 1, 2, 3, 4, 5, 6, 7, 8, 9]}$ \\
& $L=11$ & $K=6$ & $\mathbf{[0, 1, 2, 3, 4, 5, 6, 7, 8, 9, 10]}$ & $\mathbf{[0, 1, 2, 3, 4, 5, 6, 7, 8, 9, 10]}$ \\
& $L=12$ & $K=6$ & $[1, 2, 3, 4, 5, 6, 7, 11, 10, 9, 0, 8]$ & $\mathbf{[0, 1, 2, 3, 4, 5, 6, 7, 8, 9, 10, 11]}$ \\
& $L=13$ & $K=6$ & $[0, 1, 2, 3, 12, 11, 10, 4, 5, 6, 7, 8, 9]$ & $[8, 9, 0, 1, 2, 3, 4, 10, 11, 12, 5, 6, 7]$ \\
\midrule

\multirow{1}{*}{\textsc{Prod}}
& $L=10$ & $K=6$
  & $\mathbf{[0, 1, 2, 3, 4, 5, 6, 7, 8, 9]}$ &
    $\mathbf{[0, 1, 2, 3, 4, 5, 6, 7, 8, 9]}$ \\
\bottomrule
\end{tabular}
}
\end{table*}

\begin{table}[h]
\centering
\caption{Detailed runtime analysis including total exploration time and the proportion of time spent on model training within the exploration phase. For non-\textsc{Prod} order-sensitive tasks, representative timings are measured on \textsc{ReLU}, since runtime is almost identical across tasks under the shared training setup.}
\label{tab:runtime_analysis_appendix}
\begin{tabular}{lrrrrr}
\toprule
\multirow{2}{*}{\textbf{Search Params}} & \multicolumn{3}{c}{\textbf{Exploration Time [sec]}} & \multicolumn{1}{c}{\textbf{Training Ratio}} & \multirow{2}{*}{\textbf{Final Training [sec]}} \\
\cmidrule(lr){2-4} \cmidrule(lr){5-5}
 & \textbf{Total} & \textbf{Global} & \textbf{Local} & \multicolumn{1}{c}{\textbf{in Exploration (\%)}} & \\
\midrule
\multicolumn{6}{l}{\textit{\textbf{Prod Task}}} \\
$K=6, L=10$ & 11,729 & 7,001 & 4,448 & 67.6 & 28,213 \\
$K=6, L=12$ & 13,436 & 7,301 & 5,935 & 67.9 & 30,695 \\
$K=7, L=20$ & 35,981 & 35,981 & ---   & 11.7 & 38,369 \\
\midrule
\multicolumn{6}{l}{\textit{\textbf{Other (Order-sensitive) Tasks}}} \\
$K=6, L=10$ & 4,424 & 3,912 & 512 & 14.8 & 859 \\
$K=6, L=13$ & 4,880 & 4,112 & 768 & 15.1 & 899 \\
$K=7, L=20$ & 29,212 & 29,212 & --- & 1.7 & 998 \\
$K=7, L=30$ & 36,812 & 33,812 & --- & 1.7 & 1,206 \\
$K=7, L=40$ & 40,351 & 37,351 & --- & 1.9 & 1,365 \\
\bottomrule
\end{tabular}
\end{table}

\FloatBarrier
\section{Details on delay network variants}
\label{app:delay-variants-defs}

In \cref{sec:exp-delay-variants} we evaluated the loss-profiling pipeline on six delay network variants.
This appendix gives the full transition rules for each variant, the candidate-order definitions, and notes two further variants that we attempted but excluded from the main table.

\subsection{Common protocol}
For all variants, the state at time $t$ is $s(t) = (a(t), b(t), c(t), d(t))$, the input is $(s(1), s(2))$ with each component sampled uniformly from $[-25, 25]$, and the target is the time-major flattening of $s(3), \ldots, s(T)$, which has length $4(T-2)$.
Training and test samples are generated with seeds $42$ and $1337$, respectively, with $100{,}000$ training and $1{,}000$ test samples per variant.
Inputs and targets are tokenized as space-separated integers; loss is computed only on the target tokens and the end-of-sequence token.
Loss profiling uses a 1-layer GPT-2 with embedding dimension $128$ and one attention head, trained for one epoch with batch size $32$ and learning rate $5 \times 10^{-4}$, applying each candidate permutation in a per-sample round-robin during training and ranking candidates by validation loss.
Retraining uses a 6-layer GPT-2 with the same embedding dimension and one head, trained for $10$ epochs with batch size $128$ and the same learning rate, with the target order fixed throughout.

\subsection{Transition rules}

\vspace{1mm}
\noindent\textbf{Original \textsc{ReLU}.}
\[
\begin{aligned}
a(t+2) &= \operatorname{ReLU}(b(t+1)+d(t)), &
b(t+2) &= a(t+1)-c(t+1), \\
c(t+2) &= b(t)+d(t+1), &
d(t+2) &= \operatorname{ReLU}(a(t+1)+c(t)).
\end{aligned}
\]

\vspace{1mm}
\noindent\textbf{\textsc{Hidden-cause}.}
With $h(1)=h(2)=0$,
\[
\begin{aligned}
h(t+2) &= \operatorname{ReLU}(a(t+1)-d(t)), &
a(t+2) &= \operatorname{ReLU}(b(t+1)+h(t+1)), \\
b(t+2) &= a(t+1)-c(t+1), &
c(t+2) &= b(t)+d(t+1), \\
d(t+2) &= \operatorname{ReLU}(h(t+2)+c(t)). & &
\end{aligned}
\]
The latent variable $h$ is not exposed to the model; only $(a,b,c,d)$ appear in the target.

\vspace{1mm}
\noindent\textbf{\textsc{Coupled ring}.}
\[
\begin{aligned}
a(t+2) &= \operatorname{ReLU}(b(t+1)-d(t)), &
b(t+2) &= \operatorname{ReLU}(c(t+1)-a(t)), \\
c(t+2) &= \operatorname{ReLU}(d(t+1)-b(t)), &
d(t+2) &= \operatorname{ReLU}(a(t+1)-c(t)).
\end{aligned}
\]

\vspace{1mm}
\noindent\textbf{\textsc{Modular}.}
With all operations taken modulo $19$,
\[
\begin{aligned}
a(t+2) &= (b(t+1)+d(t)) \bmod 19, &
b(t+2) &= (a(t+1)-c(t+1)) \bmod 19, \\
c(t+2) &= (b(t)\cdot d(t+1)) \bmod 19, &
d(t+2) &= (a(t+1)+c(t)) \bmod 19.
\end{aligned}
\]

\vspace{1mm}
\noindent\textbf{\textsc{Sine}.}
Define $\operatorname{sine}_{32}(z) = \big(\lfloor 15\sin(2\pi z / 10)\rfloor + 16\big) \bmod 32$. Then
\[
\begin{aligned}
a(t+2) &= \operatorname{sine}_{32}(b(t+1)+d(t)), &
b(t+2) &= \operatorname{sine}_{32}(a(t+1)-c(t+1)), \\
c(t+2) &= \operatorname{sine}_{32}(b(t)+d(t+1)), &
d(t+2) &= \operatorname{sine}_{32}(a(t+1)+c(t)).
\end{aligned}
\]

\vspace{1mm}
\noindent\textbf{\textsc{Mixed nonlinear}.}
Define $\operatorname{sq}_{19}(x) = (x^{2} \bmod 19) - 9$ and $\operatorname{mod}_{19}(x) = x \bmod 19$. Then
\[
\begin{aligned}
a(t+2) &= \operatorname{ReLU}\!\big(b(t+1)-d(t)+\operatorname{sq}_{19}(c(t))\big), \\
b(t+2) &= \operatorname{mod}_{19}\!\big(a(t+1)^{2} + c(t+1) - d(t)\big), \\
c(t+2) &= \operatorname{sq}_{19}(b(t)+d(t+1)) - \operatorname{ReLU}(a(t)), \\
d(t+2) &= \operatorname{ReLU}\!\big(a(t+1)+c(t)-\operatorname{mod}_{19}(b(t+1))\big).
\end{aligned}
\]

\subsection{Candidate orders}
Each variant induces its own dependency graph, so the actual permutations associated with a candidate name are task-specific.
The five candidate names are kept consistent across variants:
\begin{itemize}[leftmargin=1.4em, labelsep=0.4em, itemsep=2pt, topsep=2pt]
    \item \texttt{time\_first}: the natural forward flattening $(a(3),b(3),c(3),d(3),\ldots,a(T),b(T),c(T),d(T))$.
    \item \texttt{variable\_first\_topological}: a topological order of the per-task dependency graph with tie-breaking $a > b > c > d$. Components from different time steps may be interleaved when the dependency graph permits.
    \item \texttt{dependency\_sorted}: keeps the time-major structure but reorders within each time step according to a task-specific within-time order (e.g., $(b,a,d,c)$ for the variants used here).
    \item \texttt{causal\_depth}: keeps the time-major structure but uses a depth-based within-time order ($(a,d,b,c)$ for the variants used here).
    \item \texttt{reverse}: complete reversal of the target sequence.
\end{itemize}
The exact within-time orderings used for each variant are stored in the per-task candidate files \texttt{T=11\_candidates.json}.

\subsection{Per-candidate validation losses from loss profiling}
\label{app:delay-profile-losses}
\Cref{tab:delay-profile-losses} reports the validation losses obtained at the end of the 1-epoch profiling phase (1-layer GPT-2, $d_{\mathrm{emb}}=128$) for each of the five candidate orders.
These are the underlying numbers behind the rank assignments shown in \cref{tab:delay-variants-summary}.
Two patterns referenced in \cref{sec:exp-delay-variants} can be read directly from this table:
on \textsc{Modular}, all five candidates lie within a $0.02$-loss band, which is the saturation/tie issue;
on \textsc{Mixed nonlinear}, the rank-1 (\texttt{variable\_first\_topological}, $2.46$) and rank-2 (\texttt{dependency\_sorted}, $2.48$) candidates are nearly indistinguishable at the early stage despite substantially different retrain accuracies ($1.4\%$ vs.\ $64.5\%$).
The \textsc{Asymmetric} row provides the underlying losses for the variant excluded in \cref{app:delay-variants-defs} (all five candidates retrain to $0\%$ regardless of profile rank).

\begin{table}[h]
\centering
\small
\setlength{\tabcolsep}{4pt}
\caption{Validation loss after the 1-epoch profiling run (1-layer GPT-2, $d_{\mathrm{emb}}=128$, $T=11$) for each of the five candidate orders on each delay network variant. \textbf{Bold} marks the rank-1 (lowest-loss) candidate per task and matches the bold entries in \cref{tab:delay-variants-summary}. The bottom row reports the \textsc{Asymmetric square-mod-ReLU} variant excluded from the main comparison.}
\label{tab:delay-profile-losses}
\begin{tabular}{lccccc}
\toprule
\textbf{Task} & \texttt{time\_first} & \texttt{var\_first\_topo} & \texttt{dep\_sorted} & \texttt{causal\_depth} & \texttt{reverse} \\
\midrule
\textsc{ReLU}            & 3.634          & \textbf{3.492} & 3.507          & 3.564          & 3.702 \\
\textsc{Hidden-cause}    & 2.706          & \textbf{2.659} & 2.710          & 2.748          & 3.067 \\
\textsc{Coupled ring}    & \textbf{0.842} & 1.266          & 0.879          & 0.842          & 1.167 \\
\textsc{Modular}         & 2.843          & 2.829          & \textbf{2.824} & 2.829          & 2.839 \\
\textsc{Sine}            & 1.441          & 1.460          & \textbf{1.320} & 1.492          & 1.788 \\
\textsc{Mixed nonlinear} & 2.479          & \textbf{2.458} & 2.475          & 2.560          & 2.515 \\
\midrule
\textsc{Asymmetric}      & 3.471          & \textbf{3.392} & 3.401          & 3.437          & 3.526 \\
\bottomrule
\end{tabular}
\end{table}

\subsection{Scaling with the number of time steps on the original \textsc{ReLU} delay network}
\label{app:delay-original-scaling}
We extend the \textsc{ReLU} delay network to larger numbers of time steps, $T \in \{11,13,15,17\}$, with corresponding target lengths $4(T-2) \in \{36, 44, 52, 60\}$.
The order picked by loss profiling, \texttt{variable\_first\_topological}, retains $\geq 94.5\%$ accuracy throughout, whereas the natural \texttt{time\_first} order degrades from $73.3\%$ to $0.3\%$, and \texttt{reverse} stays at zero (\cref{tab:delay-original-scaling}).
Together with the variant comparison in \cref{sec:exp-delay-variants}, this indicates that loss profiling identifies a learning-friendly order in delay networks as the recurrence becomes more complex and $T$ becomes larger.

\begin{table}[h]
\centering
\small
\setlength{\tabcolsep}{6pt}
\caption{Success rate (\%) on the \textsc{ReLU} delay network at varying $T$. The order picked by loss profiling, \texttt{variable\_first\_topological}, remains effective at all tested $T$ values, while \texttt{time\_first} degrades sharply.}
\label{tab:delay-original-scaling}
\begin{tabular}{lcccc}
\toprule
\textbf{Target order} & $T=11$ & $T=13$ & $T=15$ & $T=17$ \\
\midrule
\texttt{variable\_first\_topological} & \textbf{100.0} & \textbf{99.6} & \textbf{96.4} & \textbf{94.5} \\
\texttt{dependency\_sorted}           & 84.6  & 75.9  & 70.5  & 46.3 \\
\texttt{causal\_depth}                & 63.0  & 65.4  & 61.7  & 39.3 \\
\texttt{time\_first}                  & 73.3  & 58.5  & 51.8  & 0.3  \\
\texttt{reverse}                      & 0.0   & 0.0   & 0.0   & 0.0  \\
\bottomrule
\end{tabular}
\end{table}

\subsection{Variants excluded from the main table}
We additionally attempted two variants that are excluded from \cref{tab:delay-variants-summary} for the following reasons.
The \textsc{Square} variant, which applies $\operatorname{sq}_{19}$ componentwise to the original \textsc{ReLU} couplings, did not complete 10-epoch retraining within our compute budget; loss profiling assigned rank 1 to \texttt{dependency\_sorted}, but final retraining figures are unavailable.
The \textsc{Asymmetric square-mod-ReLU} variant combines mixed nonlinearities with asymmetric coupling, and the success rate stayed at $0.0\%$ for all five candidate orders under the same training budget.
A likely cause is the limited capacity of the retraining model on this task; we therefore exclude it from the comparison.

\section{Global search on delay network variants}
\label{app:delay-global-search}

In \cref{sec:exp-delay-variants} we summarize the application of the full global stage of \cref{sec:proposed-method} to delay network variants.
This appendix describes the setup, gives the per-task and per-initialization results, and lists the discovered orders.

\subsection{Setup}
\label{app:delay-global-search-setup}

The global stage is run independently for each delay-network variant on the same training set used in \cref{sec:exp-delay-variants}.
Two initialization schemes are compared:
\begin{itemize}[leftmargin=1.4em, labelsep=0.4em, itemsep=2pt, topsep=2pt]
    \item \texttt{topo} (\texttt{topological\_random}): $100$ random topological sorts of the variant's dependency graph, with random tie-breaking. This scheme requires the dependency graph to be known.
    \item \texttt{vars} (\texttt{variable\_order\_24}): the $24$ permutations of $(a,b,c,d)$ enumerated as within-time orderings, keeping the time-major layout fixed. This scheme requires no graph information.
\end{itemize}
For both schemes, the global stage uses $1$-layer GPT-2 models with one attention head, trained for one epoch with batch size $32$ at learning rate $5 \times 10^{-4}$.
The embedding dimension is $128$ in most runs and $512$ in a small number of runs where the smaller capacity yields degenerate losses (notably \textsc{Square} under \texttt{topo} at $K{=}2$).
Hierarchical depth $K$ is searched over $\{2,3,4,5\}$, and for each variant and initialization, we select the configuration whose ranked top order has the lowest validation loss as the global stage output.
The local stage is omitted because the target length $36$ is larger than the block-level refinement budget used elsewhere in the paper.
Retraining of the discovered order uses the same 6-layer GPT-2 protocol as in \cref{sec:exp-delay-variants}: 10 epochs, batch size $128$, learning rate $5 \times 10^{-4}$, evaluated on $1{,}000$ held-out samples.

\subsection{Per-task and per-initialization results}
\label{app:delay-global-search-detailed}

\Cref{tab:delay-global-search-detailed} reports, for each variant and each initialization, the global-stage validation loss of the chosen order, the success rate after fixed-order retraining, and, for reference, the success rates of \texttt{reverse} and the best of the five hand-designed candidates.

\begin{table}[h]
\centering
\small
\setlength{\tabcolsep}{4pt}
\caption{Detailed results of the global search on delay network variants with $T=11$. ``Global loss'' is the validation loss of the chosen order at the end of the global stage. ``Discovered'' is the success rate after retraining the chosen order with the 6-layer GPT-2 protocol. ``Reverse'' and ``Best of 5'' are reproduced from \cref{tab:delay-variants-summary} for reference; ``\textemdash'' indicates that the corresponding hand-designed comparison was inconclusive (see \cref{app:delay-variants-defs}).}
\label{tab:delay-global-search-detailed}
\begin{tabular}{llcccc}
\toprule
\textbf{Task} & \textbf{Init} & \textbf{Global loss} & \textbf{Discovered (\%)} & \textbf{Reverse (\%)} & \textbf{Best of 5 (\%)} \\
\midrule
\multirow{2}{*}{\textsc{Square}}              & \texttt{topo} & 2.117 & 100.0 & --- & --- \\
                                              & \texttt{vars} & 2.252 & 100.0 & --- & --- \\
\midrule
\multirow{2}{*}{\textsc{Hidden-cause}}        & \texttt{topo} & 5.418 & 69.1  & 0.5  & 96.3 \\
                                              & \texttt{vars} & 5.447 & 60.4  & 0.5  & 96.3 \\
\midrule
\multirow{2}{*}{\textsc{Coupled ring}}        & \texttt{topo} & 4.221 & 99.1  & 32.6 & 99.9 \\
                                              & \texttt{vars} & 4.298 & 99.8  & 32.6 & 99.9 \\
\midrule
\multirow{2}{*}{\textsc{Modular}}             & \texttt{topo} & 2.901 & 0.0   & 0.0  & 100.0 \\
                                              & \texttt{vars} & 2.890 & 0.0   & 0.0  & 100.0 \\
\midrule
\multirow{2}{*}{\textsc{Sine}}                & \texttt{topo} & 1.731 & 99.3  & 0.0  & 100.0 \\
                                              & \texttt{vars} & 1.740 & 100.0 & 0.0  & 100.0 \\
\midrule
\multirow{2}{*}{\textsc{Mixed nonlinear}}     & \texttt{topo} & 4.859 & 76.1  & 0.0  & 64.5 \\
                                              & \texttt{vars} & 4.890 & 6.6   & 0.0  & 64.5 \\
\midrule
\multirow{2}{*}{\textsc{Asymmetric}}          & \texttt{topo} & 5.162 & 0.0   & 0.0  & 0.0 \\
                                              & \texttt{vars} & 5.152 & 0.0   & 0.0  & 0.0 \\
\bottomrule
\end{tabular}
\end{table}

\textsc{Square} is included for completeness: the global search discovers orders that retrain to $100\%$, but the corresponding hand-designed comparison did not complete in our setup, so it is reported here rather than in \cref{tab:delay-variants-global-search}.
\textsc{Asymmetric} is the same task that was excluded from \cref{tab:delay-variants-summary} due to model capacity limits; both global search and the hand-designed candidates yield $0\%$ accuracy in our setup.

A few patterns are visible in \cref{tab:delay-global-search-detailed}.
First, on tasks where global loss is well separated from chance (\textsc{Square}, \textsc{Sine}, \textsc{Coupled ring}), retraining is highly successful regardless of initialization.
Second, the failure on \textsc{Modular} is a useful negative example: the picked-order global loss ($2.890$ and $2.901$) is in the same range as the hand-designed candidates' early-stage losses, so loss profiling cannot distinguish the chosen order from the $100\%$ candidates within the first epoch.
Third, on \textsc{Mixed nonlinear} the two initializations diverge sharply ($76.1\%$ vs.\ $6.6\%$), suggesting that the graph-based \texttt{topo} initialization is needed to access the high-accuracy basin on this task.

\subsection{Discovered orders}
\label{app:delay-global-search-orders}

We list the orders discovered by the global stage for each variant and initialization.
The input prefix $s(1), s(2)$ is unchanged; only the target tokens $s(3), \ldots, s(11)$ (length $36$) are reordered.
Each order is read row by row, in groups of nine target tokens.

\vspace{1mm}
\noindent\textbf{\textsc{Square}.}
\texttt{topo} (global loss $2.117$, retrain accuracy $100\%$):
{\small\begin{verbatim}
a(3), d(3), b(3), c(4), a(4), d(4), b(5), c(3), c(5)
a(6), d(5), b(4), a(5), c(6), b(6), d(7), a(7), d(8)
c(8), d(6), b(7), a(8), c(9), b(9), a(10), d(11), c(7)
d(9), b(8), a(9), b(10), d(10), a(11), c(11), c(10), b(11)
\end{verbatim}}
\texttt{vars} (global loss $2.252$, retrain accuracy $100\%$):
{\small\begin{verbatim}
c(3), a(3), b(3), d(3), c(4), a(4), b(4), d(4), c(5)
a(5), b(5), d(5), c(6), a(6), b(6), d(6), c(7), a(7)
b(7), d(7), c(8), a(8), b(8), d(8), c(9), a(9), b(9)
d(9), c(10), a(10), b(10), d(10), c(11), a(11), b(11), d(11)
\end{verbatim}}

\vspace{1mm}
\noindent\textbf{\textsc{Hidden-cause}.}
\texttt{topo} (global loss $5.418$, retrain accuracy $69.1\%$):
{\small\begin{verbatim}
d(3), c(4), a(3), b(3), a(4), b(5), a(6), c(3), d(5)
d(4), b(4), c(5), d(7), a(5), c(6), b(6), c(8), d(6)
b(7), a(8), c(7), d(9), b(9), a(10), a(7), b(8), a(9)
d(8), c(10), c(9), d(11), b(11), d(10), b(10), c(11), a(11)
\end{verbatim}}
\texttt{vars} (global loss $5.447$, retrain accuracy $60.4\%$):
{\small\begin{verbatim}
a(3), d(3), c(3), b(3), a(4), d(4), c(4), b(4), a(5)
d(5), c(5), b(5), a(6), d(6), c(6), b(6), a(7), d(7)
c(7), b(7), a(8), d(8), c(8), b(8), a(9), d(9), c(9)
b(9), a(10), d(10), c(10), b(10), a(11), d(11), c(11), b(11)
\end{verbatim}}

\vspace{1mm}
\noindent\textbf{\textsc{Coupled ring}.}
\texttt{topo} (global loss $4.221$, retrain accuracy $99.1\%$):
{\small\begin{verbatim}
a(3), d(3), b(3), c(4), d(4), c(5), b(5), a(6), d(7)
a(4), c(3), b(6), d(5), a(7), b(4), a(5), d(6), c(6)
d(8), c(8), c(7), b(9), b(7), a(8), d(9), c(9), b(10)
a(11), b(8), a(9), d(10), c(11), c(10), b(11), a(10), d(11)
\end{verbatim}}
\texttt{vars} (global loss $4.298$, retrain accuracy $99.8\%$):
{\small\begin{verbatim}
a(3), c(3), d(3), b(3), a(4), c(4), d(4), b(4), a(5)
c(5), d(5), b(5), a(6), c(6), d(6), b(6), a(7), c(7)
d(7), b(7), a(8), c(8), d(8), b(8), a(9), c(9), d(9)
b(9), a(10), c(10), d(10), b(10), a(11), c(11), d(11), b(11)
\end{verbatim}}

\vspace{1mm}
\noindent\textbf{\textsc{Modular}.}
\texttt{topo} (global loss $2.901$, retrain accuracy $0.0\%$):
{\small\begin{verbatim}
c(3), d(3), a(3), c(4), d(4), b(4), a(5), d(6), b(3)
c(5), a(4), b(5), c(7), b(6), a(6), d(5), d(7), c(8)
a(8), d(9), c(10), b(9), c(11), a(11), a(10), b(11), d(11)
a(7), b(8), c(6), d(8), a(9), b(7), c(9), d(10), b(10)
\end{verbatim}}
\texttt{vars} (global loss $2.890$, retrain accuracy $0.0\%$):
{\small\begin{verbatim}
c(3), a(3), b(3), d(3), c(4), a(4), b(4), d(4), c(5)
a(5), b(5), d(5), c(6), a(6), b(6), d(6), c(7), a(7)
b(7), d(7), c(8), a(8), b(8), d(8), c(9), a(9), b(9)
d(9), c(10), a(10), b(10), d(10), c(11), a(11), b(11), d(11)
\end{verbatim}}

\vspace{1mm}
\noindent\textbf{\textsc{Sine}.}
\texttt{topo} (global loss $1.731$, retrain accuracy $99.3\%$):
{\small\begin{verbatim}
a(3), c(3), d(3), d(4), b(3), c(4), a(4), b(5), a(6)
d(5), c(5), d(7), b(4), a(5), d(6), c(7), c(6), b(6)
a(7), d(8), b(7), a(8), c(9), d(9), c(8), b(9), b(8)
c(10), a(10), b(11), d(11), a(9), d(10), c(11), b(10), a(11)
\end{verbatim}}
\texttt{vars} (global loss $1.740$, retrain accuracy $100.0\%$):
{\small\begin{verbatim}
b(3), d(3), a(3), c(3), b(4), d(4), a(4), c(4), b(5)
d(5), a(5), c(5), b(6), d(6), a(6), c(6), b(7), d(7)
a(7), c(7), b(8), d(8), a(8), c(8), b(9), d(9), a(9)
c(9), b(10), d(10), a(10), c(10), b(11), d(11), a(11), c(11)
\end{verbatim}}

\vspace{1mm}
\noindent\textbf{\textsc{Mixed nonlinear}.}
\texttt{topo} (global loss $4.859$, retrain accuracy $76.1\%$):
{\small\begin{verbatim}
c(3), b(3), d(3), a(3), a(4), c(4), b(5), b(4), d(4)
d(5), a(5), d(6), c(6), c(7), c(5), a(6), b(6), a(7)
b(8), b(7), a(8), d(9), c(10), d(7), c(8), b(9), a(9)
d(10), d(8), c(9), a(10), c(11), b(11), b(10), a(11), d(11)
\end{verbatim}}
\texttt{vars} (global loss $4.890$, retrain accuracy $6.6\%$):
{\small\begin{verbatim}
a(3), d(3), c(3), b(3), a(4), d(4), c(4), b(4), a(5)
d(5), c(5), b(5), a(6), d(6), c(6), b(6), a(7), d(7)
c(7), b(7), a(8), d(8), c(8), b(8), a(9), d(9), c(9)
b(9), a(10), d(10), c(10), b(10), a(11), d(11), c(11), b(11)
\end{verbatim}}

\vspace{1mm}
\noindent\textbf{\textsc{Asymmetric}.}
\texttt{topo} (global loss $5.162$, retrain accuracy $0.0\%$):
{\small\begin{verbatim}
a(3), b(3), a(4), d(3), c(3), b(4), c(4), d(4), a(5)
d(5), c(5), b(5), a(6), b(6), c(6), a(7), c(7), c(8)
d(6), b(8), b(7), c(9), a(8), d(7), c(10), b(9), a(9)
d(8), a(10), c(11), b(10), d(9), a(11), d(10), b(11), d(11)
\end{verbatim}}
\texttt{vars} (global loss $5.152$, retrain accuracy $0.0\%$):
{\small\begin{verbatim}
d(3), a(3), c(3), b(3), d(4), a(4), c(4), b(4), d(5)
a(5), c(5), b(5), d(6), a(6), c(6), b(6), d(7), a(7)
c(7), b(7), d(8), a(8), c(8), b(8), d(9), a(9), c(9)
b(9), d(10), a(10), c(10), b(10), d(11), a(11), c(11), b(11)
\end{verbatim}}

\end{document}